# From Rainfall to Road-Network Functionality: Transferable Graph Forecasting for Flood-Disrupted Transportation Networks

**Junyan Wang[1] & Naiyu Wang[1]✉**

[1] College of Civil Engineering and Architecture, Zhejiang University, Hangzhou 310058, China

✉Corresponding: naiyuwang@zju.edu.cn

**ABSTRACT**: Extreme rainfall can rapidly degrade road-network functionality (RNF), disrupting emergency mobility, accessibility, evacuation, logistics, and infrastructure restoration. Yet existing flood-transportation impact models are often difficult to deploy across regions because they rely on local calibration, computationally intensive simulation workflows, or target-region retraining. This study proposes **G-R2R**, a transferable spatiotemporal graph-learning framework that reframes rainfall-induced road-network disruption from a site-specific surrogate prediction task into a region-conditioned, topology-flexible graph forecasting problem for unseen transportation networks. The framework integrates ego-centered subgraph sampling, terrain-conditioned graph representation, region-conditioned dual-level static encoding, and dynamic rainfall–network feature learning. Static terrain, land-cover, drainage-proximity, edge-vulnerability, and network-performance attributes, together with the terrain-conditioned graph structure, are fused with rainfall forcing and recent RNF states to forecast future node-level road-network functionality across heterogeneous road networks. G-R2R is developed and evaluated using simulation-derived RNF sequences from ten flood-prone regions in Wenzhou, China, under unseen-region testing protocols. Comparative baseline evaluation shows that G-R2R consistently outperforms representative temporal and spatiotemporal graph forecasting models, achieving a stabilized MAPE of 2.33% and an $R^2$ of 0.855 under the fixed unseen-region benchmark. Leave-one-area-out evaluation further shows consistent cross-region generalization, with stabilized MAPE values ranging from 2.30% to 3.54%. Ablation analysis supports the contributions of the terrain-conditioned graph representation and region-conditioned dual-level static encoding. A historical-event forecast replay of Typhoon Fitow further shows that G-R2R reproduces the principal temporal trajectories and node-level spatial patterns of the simulation-derived RNF reference under observed rainfall forcing without target-region retraining. These results support rapid and transferable node-level RNF forecasting across the regional road-network domains represented in this study.

## 1. INTRODUCTION

Extreme rainfall is not only a hydrologic hazard, it is also a real-time transportation-functionality problem. When intense rainfall overwhelms urban drainage and road infrastructure, disruption does not remain confined to inundated road segments. It propagates through network connectivity, reduces accessibility, delays emergency response, constrains evacuation and logistics, and slows the restoration of interdependent infrastructure systems. As heavy precipitation becomes more frequent and intense under a changing climate (Calvin et al., 2023), transportation agencies increasingly need forecasts that can anticipate not only where flooding may occur, but also how road-network functionality (RNF)—the ability of a road network to support connectivity, mobility, accessibility, and emergency serviceability—will degrade and recover as rainfall evolves.

This operational need exposes a central limitation in current flood-transportation modeling: many models are accurate only where they are calibrated. A model that must be rebuilt, recalibrated, or retrained for each city or region cannot meet the geographic scale, time pressure, and data scarcity of emergency response. This limitation is especially important for rainfall-induced road disruption, where historical impact records are sparse, traffic sensors may be incomplete, drainage data are often unavailable or restricted, and flood-response mechanisms vary across terrain, land-cover, drainage context, and road-network topology. The key challenge is therefore not only to improve prediction accuracy within a known network, but to learn transferable relationships between rainfall forcing, regional vulnerability, transportation topology, and functionality loss. Transferability determines whether flood-transportation intelligence can move beyond isolated case studies toward scalable operational deployment.

Existing studies have advanced flood-transportation analysis from several directions (Suarez et al., 2005; Stamos et al., 2015; Bucar and Hayeri, 2020; Yang et al., 2020), but these advances remain fragmented. Physics-based studies have established how inundation affects road accessibility, travel speed, connectivity, emergency-service access, and network resilience (Hammond et al., 2015; Pregnolato et al., 2017; Kermanshah and Derrible, 2017; Zhou et al., 2023; He et al., 2024; Mitropoulos et al., 2025; He et al., 2026a; Liu et al., 2026; He et al., 2026b). These studies provide physical realism and transportation relevance, but they often require high-resolution local data, local hydraulic or hydrodynamic modeling, region-specific parameterization, and substantial computation. Their deployment across many heterogeneous regions during fast-evolving emergencies therefore remains difficult.

Data-driven transportation forecasting offers speed and strong capability for learning spatiotemporal dependencies. Spatiotemporal graph neural networks, including DCRNN, STGCN, Graph WaveNet, and ASTGCN, have become influential for forecasting traffic flow, speed, and demand over transportation networks (Li et al., 2018; Yu et al., 2018; Wu et al., 2019; Guo et al., 2019; Wang et al., 2020). Recent studies have begun extending graph-learning approaches to flood-related transportation applications, including traffic-flow prediction under urban flooding and flood-impact assessment on highway networks (Roy et al., 2021; Rahman and Hasan, 2023; Lin et al., 2026; Dong et al., 2026). However, these methods are still largely designed for known network settings, recurrent traffic variables, or specific flood-impact assessment tasks. They do not directly address the problem of forecasting RNF degradation and recovery in previously unseen regions, where the target state is shaped by extreme rainfall, terrain, drainage exposure, road topology, and infrastructure vulnerability.

Simulation-trained surrogate models provide another pathway by using physics-based simulations to generate learning targets for rapid prediction. In urban flood modeling, surrogate approaches based on CNN, ConvLSTM, attention mechanisms, and physics-informed deep learning have shown strong potential for accelerating inundation prediction and improving spatial generalization (Zahura et al., 2020; Guo et al., 2022; Zhang et al., 2022; do Lago et al., 2023; Lou et al., 2026; Feng et al., 2026). However, these models usually predict hydrodynamic states such as flood depth, inundation extent, or flood evolution, rather than transportation-network functionality. Prior Rainfall-to-RNF surrogate modeling extends this idea toward transportation-impact forecasting by learning from simulation-derived RNF labels (Wang and Wang, 2024). Yet a site-specific surrogate often inherits the spatial specificity of its simulation environment. When applied to a new road network, it may still require new simulations, local preprocessing, or retraining. It accelerates local prediction, but does not by itself solve transferable RNF forecasting.

Meanwhile, cross-city transfer learning has become an active topic in transportation AI. Recent studies have explored distribution alignment, transferable source-region selection, graph-structure transfer, invariant spatial representation, few-shot learning, and zero-shot traffic forecasting across cities or regions (Wang et al., 2019; Tang et al., 2022; Jin et al., 2022; Liu et al., 2023; Ouyang et al., 2023; Chen et al., 2024; Yang et al., 2025; Zhang et al., 2025; Zhou and Chen, 2026; Wang et al., 2026). Recent work has also extended transfer learning toward traffic-network resilience assessment, where source-city traffic patterns are transferred to target cities and adapted through target-city fine-tuning (Cui et al., 2026). These studies show that transferability is now recognized as a major deployment bottleneck in transportation intelligence. However, most transfer-learning studies focus on routine traffic states, mobility patterns, or resilience representations derived from observed traffic indicators. Under extreme rainfall, such observations reflect both network serviceability and adaptive travel demand: flooding may degrade road capacity or connectivity, while travelers may cancel, postpone, or reroute trips. Consequently, similar observed traffic states can correspond to different levels of underlying network degradation, making observed traffic indicators alone insufficient to uniquely characterize rainfall-induced RNF loss. This ambiguity is further compounded by sparse disruption labels and traffic-sensing records in flood-prone regions.

This creates a critical gap in transportation forecasting for extreme rainfall. Existing methods can model flood impacts, forecast traffic states, accelerate flood simulations, or transfer traffic representations, but they do not yet provide a unified framework for transferable forecasting of rainfall-induced road-network functionality across heterogeneous road-network domains. Such a framework must be topology-flexible to accommodate differences in road-network size and structure, region-conditioned to account for variations in geographic and infrastructure context, and dynamically rainfall-aware to capture functionality degradation and recovery as rainfall evolves. In this context, the key transfer problem is whether rainfall–terrain–topology–functionality relationships learned from multiple source regions remain applicable to a previously unseen regional network with different network and regional characteristics under a broadly shared hydroclimatic background. Addressing this problem is an important step toward moving beyond site-specific flood-transportation surrogate modeling.

To address this gap, this study proposes the **Generalized Rainfall-to-Road Network Functionality model**, or **G-R2R**, a transferable spatiotemporal graph learning framework for rapid forecasting of transportation-network functionality under extreme rainfall. Rather than learning a separate surrogate for each road network, G-R2R learns a shared predictive mapping across multiple regional road-network domains. The framework integrates topology-flexible ego-centered subgraph sampling, terrain-conditioned graph representation, region-conditioned dual-level static encoding, and dynamic rainfall–network feature learning. Static terrain, land-cover, drainage-proximity, edge-vulnerability, and network-performance attributes, together with the terrain-conditioned graph structure, are fused with rainfall forcing and recent RNF states to forecast future node-level road-network functionality. Through this design, G-R2R is formulated to accommodate differences in road-network topology and regional context within a common forecasting model, thereby supporting application to previously unseen target networks without target-region retraining.

From a transportation-operations perspective, the key output of G-R2R is time-evolving **node-level road-network functionality**, representing the functional state of the road network itself under flood-constrained conditions. These forecasts indicate when network serviceability is likely to deteriorate, where accessibility bottlenecks may emerge, which local corridors are likely to lose redundancy, and where recovery may begin. Such information can provide network-level support for identifying potentially degraded areas, screening critical road-network locations, and informing emergency-resource allocation and recovery prioritization during rapidly evolving rainfall events. The proposed framework is evaluated across ten flood-prone regions in Wenzhou, China, under unseen-region testing protocols, complemented by ablation analysis, prediction-error analysis, and a historical-event forecast replay of Typhoon Fitow. Together, these evaluations examine whether a model trained on multiple source regions can transfer rainfall–terrain–topology–functionality relationships to previously unseen road networks with different topological and regional characteristics under a broadly shared hydroclimatic background.

The main contributions of this study are as follows:

1. **A transferable transportation-network functionality forecasting formulation** that reframes rainfall-induced road disruption from a site-specific surrogate prediction task into a graph-domain learning problem for heterogeneous and previously unseen road networks, enabling rapid assessment of node-level road-network serviceability loss and recovery without target-region retraining.
2. **A topology-flexible spatiotemporal graph learning architecture** that uses ego-centered subgraph sampling and terrain-conditioned adjacency to support RNF forecasting over road networks with heterogeneous sizes, structures, densities, and connectivity patterns.
3. **A region-conditioned dual-level static encoding strategy** that embeds node-scale and neighborhood-scale terrain, land-cover, drainage-proximity, edge-vulnerability, and network-performance information into the predictive mapping, enabling the model to account for regional differences in geographic and infrastructure context.
4. **A transportation-oriented evaluation framework** that jointly examines predictive accuracy, cross-region transferability, temporal stability, and event-level spatiotemporal functionality evolution in unseen road networks, thereby evaluating the model beyond conventional pointwise forecasting metrics.

By shifting the emphasis from site-specific flood-impact simulation toward transferable transportation-network functionality forecasting across heterogeneous road-network domains, G-R2R provides a framework for rapid assessment of rainfall-induced network functionality and its temporal evolution.

## 2. Problem Formulation: From Single-Region R2R to Transferable G-R2R Forecasting

Building on the deployment gap identified in Section 1, this section reformulates Rainfall-to-RNF prediction from a region-specific forecasting task into a transferable graph-domain learning problem. The key shift is from learning a model for one road network to learning a shared predictive mapping that can operate across heterogeneous networks and generalize to unseen regions without target-region retraining.

RNF during extreme rainfall evolves as a coupled spatiotemporal process. Its dynamics are driven by rainfall intensity, rainfall accumulation, road-network topology, and static geographic or infrastructure conditions that shape flood exposure and vulnerability. For a given region, let the road network be represented by a graph

$$G = (V, E, A) \tag{1}$$

where V denotes road nodes, $E$ denotes road segments, and $A$ denotes the weighted adjacency matrix defined over the road-network connectivity to represent spatial relationships among connected nodes.

Let $\boldsymbol{q}_t \in \mathbb{R}^{|V|}$ denote the node-level RNF vector at time t, where each entry represents the functionality of a road node in supporting connectivity, travel efficiency, and emergency serviceability. Let $\boldsymbol{p}_{t-H+1:t}$ denote the observed rainfall sequence over the past horizon $H$, and let $\hat{\boldsymbol{p}}_{t+1:t+\tau}$ be the rainfall forecast for the next τ time steps. For a single region, the R2R forecasting task can be expressed as a graph-conditioned sequence-to-sequence mapping from recent rainfall and RNF states to future RNF states. Define the historical RNF sequence as $\boldsymbol{Q}_{t\text{-}H+1:t} = [\boldsymbol{q}_{t-H+1}, \cdots, \boldsymbol{q}_t]$. A spatiotemporal encoder first maps historical RNF, observed rainfall, and forecast rainfall into a latent dynamic representation:

$$\boldsymbol{Z}_{d,t} = E_d\left(\boldsymbol{Q}_{t\text{-}H+1:t}, \boldsymbol{p}_{t\text{-}H+1:t}, \widehat{\boldsymbol{p}}_{t+1:t+\tau}; G\right) \tag{2}$$

A decoder $D_\Theta$ then maps this latent representation to the future RNF sequence:

$$\widehat{\boldsymbol{Q}}_{t+1:t+\tau} = D_\Theta\left(\boldsymbol{Z}_{d,t}; G\right) \tag{3}$$

where $\widehat{\boldsymbol{Q}}_{t+1:t+\tau}$ denotes the predicted RNF sequence. Eqs. (2)-(3) define the essential structure of single-region R2R forecasting: the model must learn temporal dependencies in rainfall and RNF evolution while accounting for spatial interactions over the road-network graph.

However, this single-region formulation is insufficient for transfer across heterogeneous road-network domains. A model trained for one road network may learn relationships tied to local terrain, drainage conditions, road geometry, network density, and regional vulnerability characteristics. When applied to a previously unseen regional network, it may encounter different graph size, topology, and static geographic and infrastructure conditions. Without explicit mechanisms for cross-region generalization, single-region R2R models may require local retraining or redevelopment when transferred to a different road-network domain.

To address this limitation, we extend R2R prediction to a generalized cross-region forecasting problem. Let $C$ denote a set of source regions. For each region $c \in C$, the road network is represented as

$$G^{(c)} = (V^{(c)}, E^{(c)}, A^{(c)}) \tag{4}$$

Let $\psi^{(c)}$ denote static geographic and infrastructure attributes associated with region $c$, including node-level terrain and land-cover descriptors, drainage-proximity indicators, edge-level topographic weights, and network-performance features. The objective is to learn a shared predictive mapping across heterogeneous source-region graphs that can be applied to a previously unseen target-region graph without target-region retraining.

The generalized formulation is expressed as:

$$\boldsymbol{Z}_s^{(c)} = E_s\left(\psi^{(c)}; G^{(c)}\right) \tag{5}$$

$$\boldsymbol{Z}_{d,t}^{(c)} = E_d\left(\boldsymbol{Q}_{t-H+1:t}^{(c)}, \boldsymbol{p}_{t-H+1:t}^{(c)}, \widehat{\boldsymbol{p}}_{t+1:t+\tau}^{(c)}; G^{(c)}\right) \tag{6}$$

$$\widehat{Q}_{t+1:t+\tau}^{(c)} = D_\Theta\left(\boldsymbol{Z}_t^{(c)}, \boldsymbol{Z}_s^{(c)}; G^{(c)}\right) \tag{7}$$

where $E_s$ is a static encoder, $E_d$ is a spatiotemporal encoder, and $D_\Theta$ is the decoder. Unlike the single-region formulation in Eqs. (2)–(3), Eqs. (5)–(7) explicitly condition prediction on both dynamic rainfall–RNF evolution and region-specific static context. The formulation therefore requires the model to learn shared relationships among rainfall forcing, network topology, regional geographic and infrastructure context, and functionality degradation.

Fig. 1 illustrates the training–deployment logic of G-R2R. The model is trained on multiple source regions using rainfall inputs, recent RNF states, road-network topology, and static geographic features. It is then applied to a previously unseen target-region graph using its available rainfall, road-network topology, and static contextual information, without target-region retraining.

This formulation leads to three modeling requirements. First, the model must support **topology-flexible graph representation**, because road networks differ in node counts, edge structures, densities, and connectivity patterns. A transferable RNF forecasting model cannot be tied to a fixed spatial grid or a single training adjacency matrix. Second, the model must incorporate **region-conditioned static representation**, because similar rainfall forcing can lead to different functionality responses depending on terrain, land-cover, drainage proximity, road elevation, and network redundancy. Third, the model must preserve dynamic **rainfall–network feature learning**, because RNF degradation and recovery depend on rainfall timing, accumulated rainfall, forecast rainfall, and recent network states.

G-R2R is designed around these requirements. Ego-centered subgraph sampling provides a topology-flexible graph representation that can be extracted from road networks with different sizes and structures. Region-conditioned static encoders embed node-level and neighborhood-level geographic and infrastructure attributes into the predictive mapping. A dynamic spatiotemporal graph module then learns how rainfall forcing and recent RNF states evolve over the terrain-conditioned road graph. Because observed extreme-event disruption records are sparse, the model is trained using simulation-derived RNF sequences generated across heterogeneous source regions. Together, these components provide a unified formulation for forecasting road-network functionality across heterogeneous regional graphs under extreme rainfall.

The formulation also defines the data and modeling needs addressed in the following sections. Section 3 constructs a simulation-derived graph dataset that provides heterogeneous source regions, simulation-derived RNF labels, and topology-flexible graph samples. Section 4 then implements the G-R2R architecture through ego-centered subgraph sampling, region-conditioned dual-level static encoding, and dynamic rainfall–network feature learning.

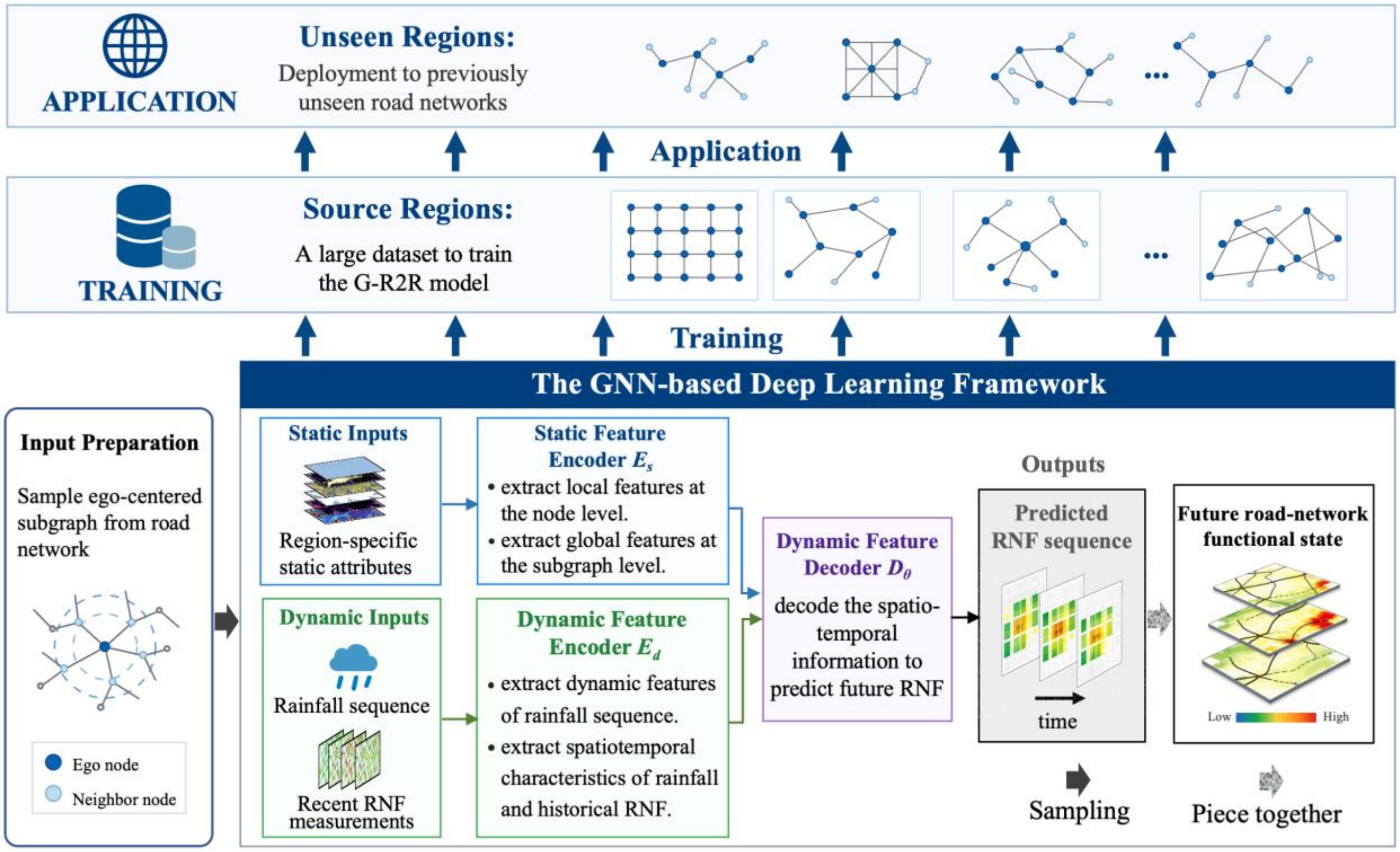

**Fig. 1. Overview of the G-R2R framework for transferable road-network functionality forecasting.** The model learns from multiple source regions and is deployed to previously unseen road networks without target-region retraining by integrating rainfall forcing, recent RNF states, static geographic features, and graph-structured transportation networks.

## 3. Study Regions and simulation-derived Transferable Graph Dataset Construction

Following the generalized formulation in Section 2, the dataset must satisfy three requirements: it must include multiple source regions with measurable variation in road-network structure and regional characteristics, provide simulation-derived RNF labels under diverse rainfall scenarios, and represent each road network through a topology-flexible graph representation. This section describes how the study areas, static geographic and network attributes, regional-scale characteristics, simulation-derived RNF labels, and graph samples are constructed to meet these requirements.

### 3.1. Study Regions and Road-Network Selection

This study uses Wenzhou, Zhejiang Province, China, as the multi-region study domain for constructing the RNF forecasting dataset and evaluating cross-region model transfer. Wenzhou is located along the East China Sea coast and is characterized by a dense river network, complex coastal–mountain terrain, and a subtropical monsoon climate. The city experiences a prolonged rainy season from May to July, followed by a typhoon season that typically extends through September. These hydroclimatic and geographic conditions make Wenzhou highly susceptible to urban flooding, waterlogging, and rainfall-induced transportation disruption.

To evaluate whether G-R2R can generalize across different regional road-network domains, we selected ten flood-prone urban communities within Wenzhou: Lingxi (LX), Yueqing (YQ), Lucheng (LC), Ouhai (OH), Sanyang (SY), Yongzhong (YZ), Aojiang (AJ), Wencheng (WC), Pingyang (PY), and Longgang (LG). These communities were delineated using hydrological boundaries, administrative maps, and regional flood-planning documents. Their geographic locations are shown in Fig. 2(a), and their road-network structures are summarized in Fig. 2(b) and Table 1.

The road networks were extracted from OpenStreetMap and preprocessed to retain arterial and major roads that serve as critical lifelines during emergency operations. Minor local streets were excluded so that the modeled networks focus on routes most relevant to evacuation, emergency supply distribution, rescue access, and post-disaster recovery. This preprocessing step ensures that the RNF forecasts target the portion of the transportation system most important for operational decision-making under flood conditions.

The selected regions vary in spatial extent and road-network structure. Their areas range from 2.84 to 21.72 km², node counts range from 72 to 403, and road-segment counts range from 106 to 637. As shown in Fig. 2(b), the communities also exhibit visibly different network layouts, from compact and sparse systems to larger and denser urban grids. Together, the ten selected communities form the multi-region study set used in the subsequent analysis. Their regional characteristics and heterogeneity are examined quantitatively in Section 3.2.

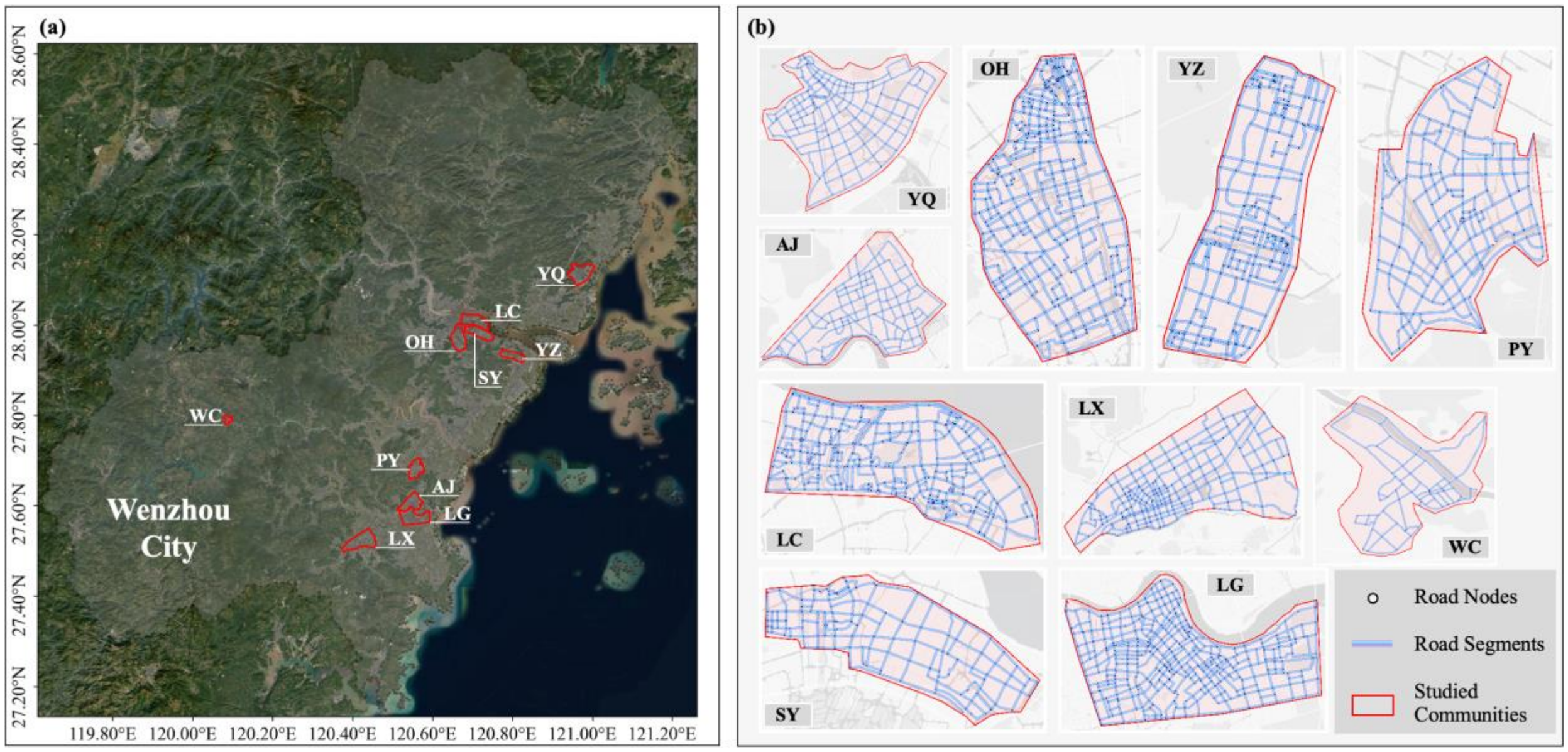

**Fig. 2. Information on the study area.** (a) Location map and (b) Road network topology diagram.

**Table 1.** Road network information for the selected study regions.

| Community | Area （km²） | No. of Nodes (n) | No. of Road Segments | Total Road Length (km) |
|---|---|---|---|---|
| **LX** | 21.72 | 233 | 382 | 122.00 |
| **LG** | 20.39 | 403 | 637 | 154.53 |
| **LC** | 16.40 | 359 | 539 | 115.28 |
| **YQ** | 20.76 | 355 | 544 | 128.50 |
| **OH** | 16.71 | 343 | 532 | 117.16 |
| **YZ** | 10.31 | 237 | 355 | 77.55 |
| **PY** | 12.55 | 174 | 271 | 75.56 |
| **AJ** | 15.02 | 126 | 196 | 71.04 |
| **WC** | 2.84 | 72 | 106 | 20.27 |
| **SY** | 14.23 | 142 | 218 | 72.69 |

### 3.2. Regional-Scale Characterization of the Study Areas

To characterize regional-scale differences among the ten study areas, four complementary descriptors were used to represent terrain morphology, river-network context, and road-network topology: relative relief (RR), TPI interquartile range (TPI-IQR), river-network density (RND), and bridge-edge fraction (BEF). RR and TPI-IQR were derived from a 30-m-resolution Digital Elevation Model (DEM) obtained from the U.S. Geological Survey (USGS). RR was calculated as the difference between the 90th and 10th percentiles of valid DEM elevations within each study area. TPI was calculated following Weiss (2001) using a 300-m reference window, and TPI-IQR was defined as the interquartile range of the resulting cell-level TPI values within each study area. RND was obtained from the national grid drainage-density dataset of Peng et al. (2024), with regional values calculated as area-weighted averages of the original grid cells intersecting each study area. The dataset represents river-centerline density in km/km² on a 1-km grid, and the original grid support was retained during regional aggregation. BEF was calculated from the preprocessed OpenStreetMap road networks introduced in Section 3.1. The data basis and definitions of the four descriptors are summarized in Table 2.

**Table 2.** Regional-scale descriptors used to characterize the ten study areas.

| Regional dimension | Descriptor | Data basis | Definition |
|---|---|---|---|
| Terrain morphology | Relative relief (RR) | 30-m USGS DEM | Difference between the 90th and 10th percentiles of valid DEM elevations within each study area |
| Terrain morphology | TPI interquartile range (TPI-IQR) | DEM-derived TPI following Weiss (2001) | Interquartile range of cell-level TPI values within each study area, representing the variability of relative topographic position |
| River-network context | River-network density (RND) | National grid drainage-density dataset of Peng et al. (2024) | Area-weighted mean of the original river-density grid cells intersecting each study area |
| Road-network topology | Bridge-edge fraction (BEF) | Preprocessed OSM road network described in Section 3.1 | Fraction of road edges identified as graph bridges, whose removal increases the number of connected components |

Between-region variation was summarized using the coefficient of variation (CV) calculated from the ten region-level values of each descriptor. The CV was additionally recalculated after excluding each study area in turn to examine the consistency of the observed regional dispersion. As shown in Fig. 3(a), the four descriptors exhibit different levels of between-region variation. TPI-IQR and RR show comparatively pronounced variation across the study areas, reflecting substantial differences in terrain morphology. BEF also varies clearly among the road networks, indicating differences in their dependence on structurally critical links. RND shows a more moderate but still discernible variation across the ten regions. The leave-one-region-out results generally retain these patterns, although the dispersion of TPI-IQR is more sensitive to the inclusion of WC.

Fig. 3(b) further illustrates the distinct regional profiles of the ten study areas across the four descriptors. For example, YQ exhibits pronounced terrain relief but relatively lower RND and BEF, whereas LG combines low terrain relief with relatively high RND and BEF. These contrasts show that the study areas differ in both the magnitude and combination of terrain, river-network, and road-network characteristics. Together with the differences in network size and layout reported in Section 3.1, these descriptors show that the ten study areas have different but partly overlapping regional profiles in terrain, river-network context, and road-network structure. Within the broadly shared hydroclimatic background described in Section 3.1, these differences represent the principal regional domain variations examined in the subsequent cross-region evaluation.

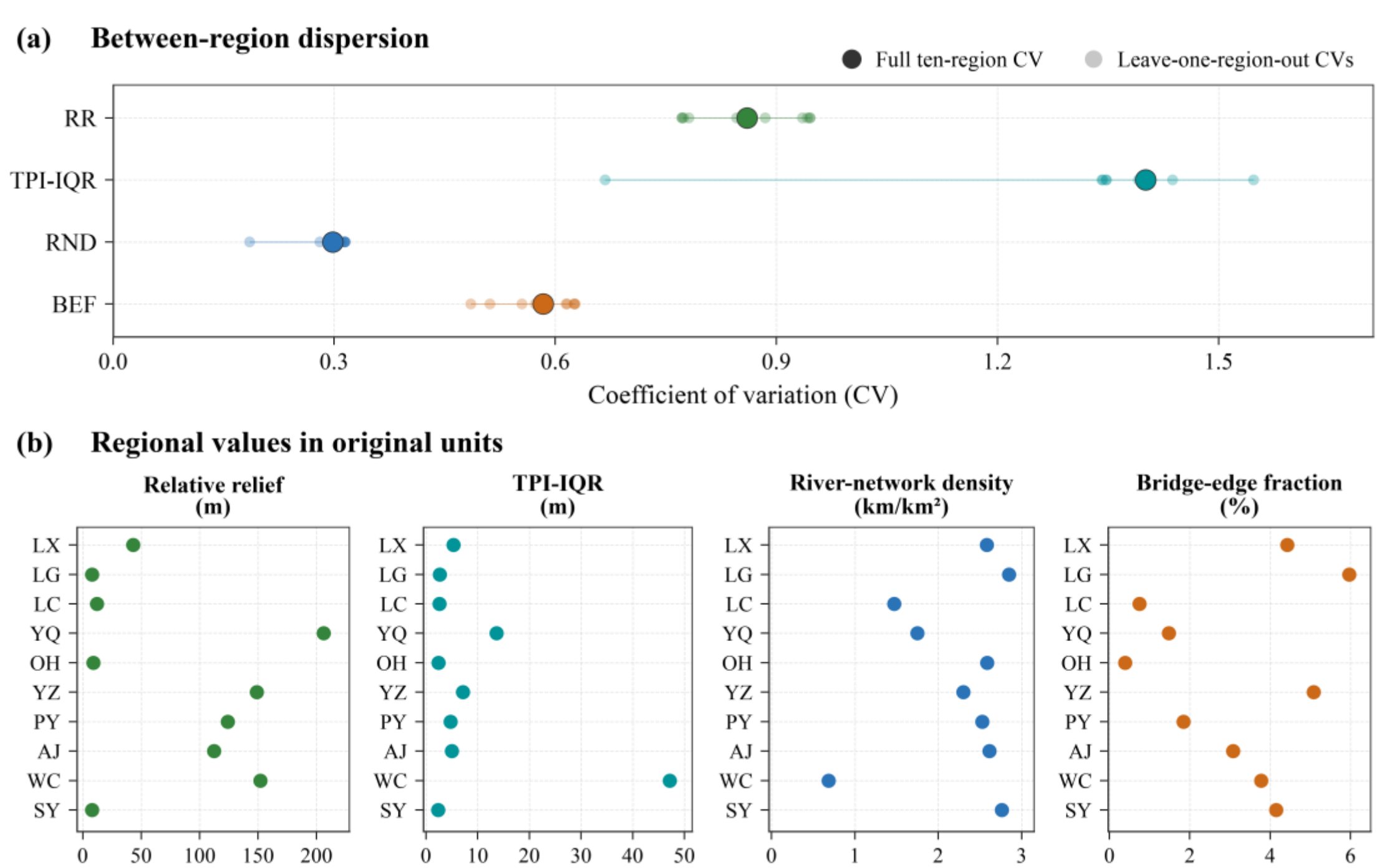

**Fig. 3. Regional-scale characterization of the ten study areas**. (a) Between-region dispersion of four regional-scale descriptors. Large filled markers denote coefficients of variation (CVs) calculated using all ten regions, while smaller markers denote CVs recalculated after excluding each region in turn. (b) Regional values of relative relief, TPI IQR, river-network density, and bridge-edge fraction in their original units. The same region order is maintained across all subpanels to facilitate comparison of regional attribute combinations.

### 3.3. Static Regional Attributes for G-R2R

Region-conditioned forecasting requires static predictive attributes that describe the local geographic, hydrological, and network conditions associated with rainfall-induced RNF degradation. G-R2R therefore incorporates four groups of static attributes: topographic features, land-cover classification, drainage-proximity information, and road-network performance features. Together, these attributes provide the local and neighborhood-scale contextual information used by the model to condition RNF prediction.

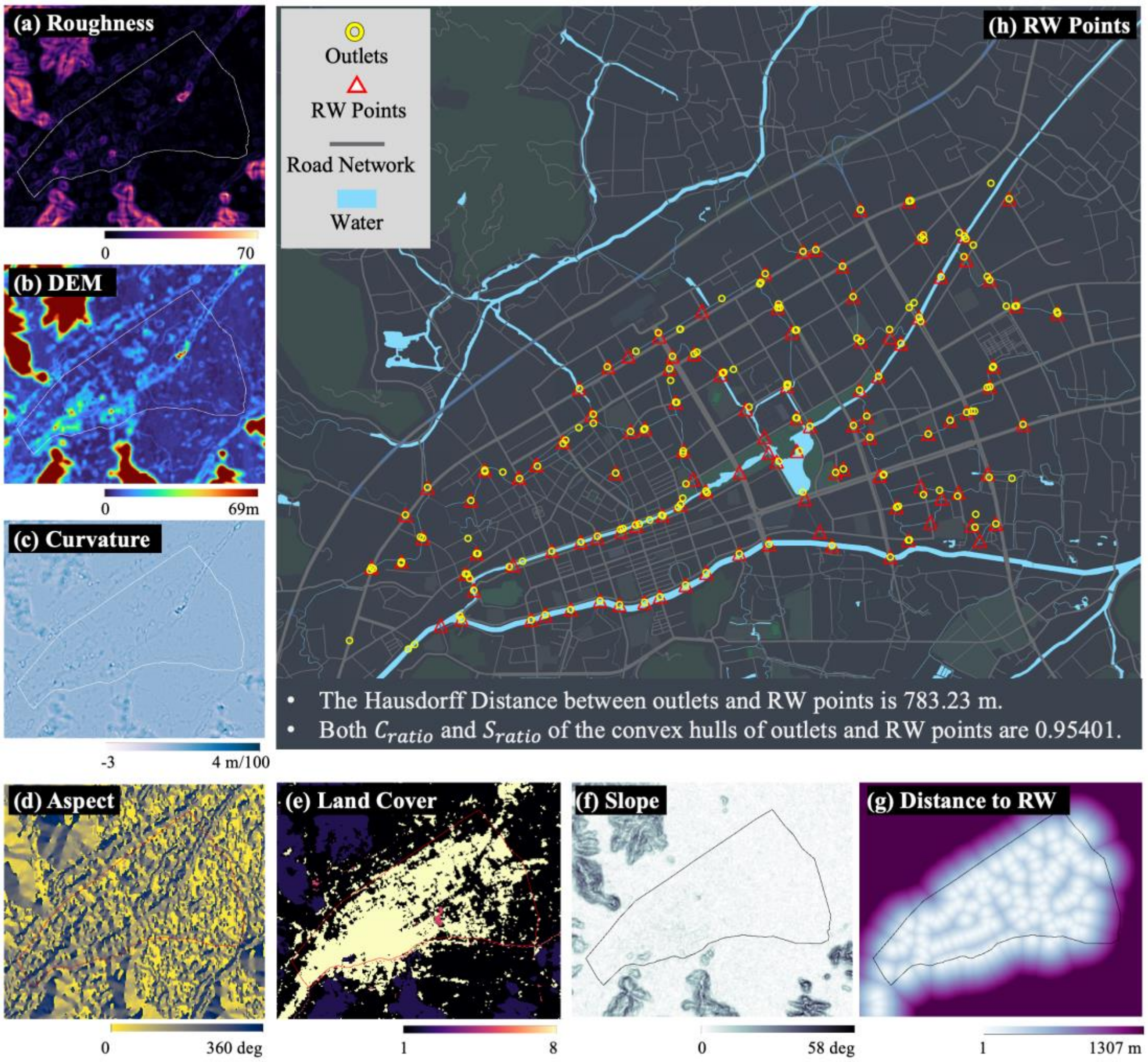

**Fig. 4.** Static geographic feature layers for the representative community LX.

**Topographic features.** The 30-m DEM introduced in Section 3.2 was used as the base terrain layer. In addition to elevation, four secondary topographic attributes—slope, curvature, aspect, and roughness—were derived from the DEM using ArcGIS. These terrain features describe local morphology associated with runoff concentration, ponding tendency, and spatial flood propagation.

**Land-cover classification.** Land-cover data were incorporated to represent spatial variations in surface characteristics. The data were obtained from the 10-m-resolution ROM-GLC global land-cover dataset and resampled to match the 30-m topographic grid. The resulting land-cover layer provides information on the spatial distribution of built-up surfaces and other land-cover types associated with localized flood response.

**Drainage-proximity indicator.** Drainage context was represented by the distance from each grid cell to the nearest drainage discharge point. Direct drainage-outlet information, however, is difficult to obtain consistently across regions. We therefore approximate potential drainage outlets using RW points, defined as intersections between the road network and natural water bodies. This proxy is based on the common configuration of urban drainage systems, in which drainage infrastructure is frequently aligned with road corridors and conveys runoff toward adjacent rivers or lakes. The spatial representativeness of the RW points was further examined against available drainage-system information, as illustrated in Fig.4(h), yielding a Hausdorff distance of 783.23 m and convex-hull similarities of 0.954 for both area and perimeter. Based on these RW points, the distance from each grid cell to its nearest RW point was calculated to construct the $D_{RW}$ raster shown in Fig. 4(g). This raster serves as a proxy for drainage proximity and is used as a neighborhood-scale static input to G-R2R

**Road-network performance features.** In addition to geographic and hydrological attributes, we include road-network performance indicators that describe the functional role and redundancy of each node under normal operating conditions. These indicators provide complementary network-level information to the preceding raster-based static geographic features. Their calculation is introduced in the subsequent RNF assessment section, while their integration with the raster-based static features in G-R2R is described in Section 4.1.

### 3.4. Simulation-Derived RNF Label Generation

This subsection describes how dynamic RNF labels are generated for training G-R2R. Because systematic observations of road-network disruption during extreme rainfall are sparse and incomplete, while observed traffic measures are also influenced by concurrent changes in travel demand, we use simulation-derived RNF as a consistent network-level learning target. RNF characterizes road-network functionality under flood-constrained conditions in terms of connectivity, travel efficiency, and pathway redundancy. As shown in the upper row of Fig. 5, the workflow consists of three steps: stochastic generation of extreme rainfall events, hydrological and hydrodynamic flood simulation, and node-level RNF assessment. The lower row of Fig. 5, which transforms the simulated outputs into graph samples for G-R2R, is discussed in Section 3.5.

**Stochastic generation of extreme rainfall events.**

Extreme rainfall events are characterized not only by total accumulation and peak intensity, but also by temporal structure, including duration, peak timing, inter-event intervals, and intra-event concentration. To represent this variability, we analyzed historical hourly rainfall records from 339 meteorological stations in Wenzhou from 1970 to 2020, following Wang and Wang (2024). Based on the empirical distributions derived from these records, we used the stochastic rainfall generator of Wu et al. (2006) to simulate 8,378 24-hour rainfall sequences at hourly resolution. These synthetic sequences include front-loaded, centrally peaked, back-loaded, and multi-modal rainfall patterns, allowing the training dataset to cover a broad range of extreme-rainfall temporal dynamics. The same rainfall-event library was used to drive the simulations for all ten study regions, providing a common rainfall-forcing basis for the multi-region dataset.

Fig. 6 summarizes the generated rainfall dataset. Panel (a) shows representative rainfall profiles with different temporal concentrations. Panel (b) presents the distributions of peak timing and event duration, while panels (c) and (d) show the ranges of peak intensity and total accumulation. Peak intensities range from 10 to 80 mm/h, and total accumulations range from 45.5 to 652.5 mm, providing diverse hazard forcings for subsequent flood and RNF simulations.

**Hydrological and hydrodynamic flood simulation.**

The generated rainfall sequences were used as inputs to physics-based flood simulations for the ten selected regions. For each community, a drainage-network model was developed using available infrastructure data. The Storm Water Management Model (SWMM) was used to estimate hourly overflow rates at catchment drainage wells (Gironás et al., 2010). These overflow outputs were then fed into a cellular automaton–based flood model to simulate flood evolution at 30-m spatial resolution (Guidolin et al., 2016). For each study region, the coupled flood-simulation model was calibrated and validated using region-specific drainage and flood-control information together with locally available inundation records before being used for scenario generation. The regional calibration and validation procedure, together with a representative validation case for Lingxi, is provided in Appendix A; further implementation details are reported in Wang and Wang (2024). This resolution is consistent with the DEM-based terrain features introduced in Section 3.2, ensuring spatial alignment among rainfall forcing, terrain attributes, inundation outputs, and road-network features.

Within this framework, detailed drainage-network information is used in the offline physics-based simulations to generate simulation-derived RNF labels and evaluation references. It is not required for G-R2R inference, for which drainage conditions are represented through the $D_{RW}$ feature described in Section 3.3.

**Assessment of road-network functionality.**

Flood inundation depths obtained from the simulations were first translated into time-varying road-segment service conditions following the RNF assessment framework of Wang and Wang (2024). Each road segment $l$ was divided into spatial units consistent with the resolution of the inundation simulation. For each unit, the simulated water depth was converted to the corresponding maximum safe travel speed using the depth-disruption relationship proposed by Pregnolato et al. (2017). These unit-level speeds were then used to calculate the flood-constrained traversal time $TTI_l^t$ of each road segment. For a passable segment, its service level at time $t$ was calculated as $s_l^t = TTI_l^0 / TTI_l^t$, where $TTI_l^0$ denotes the traversal time under normal operating conditions. A road segment was treated as impassable, with $s_l^t = 0$, when the inundation depth of any constituent unit reached 0.30 m. The depth-speed relationship therefore serves as an intermediate step in deriving flood-constrained segment serviceability. Detailed formulations of the depth-speed relationship and segment traversal time are provided in Wang and Wang (2024).

Road-segment service levels characterize local flood-induced disruption, whereas node-level road-network functionality also depends on the availability, redundancy, and service conditions of the routes connecting each node to other destinations. The segment-level states were therefore further aggregated through independent pathways (IPWs) using the Weighted Independent Pathway (WIPW) framework of Zhang and Wang (2016). A pathway between an origin-destination (O-D) pair consists of a sequence of connected road segments. Two pathways between the same O-D pair are defined as independent pathways when they do not share any road segment. For each O-D pair, the shortest pathway was first identified using Dijkstra's algorithm and treated as the first IPW. The road segments belonging to that pathway were then removed from the network, and the shortest-path search was repeated on the updated network until no further pathway existed between the two nodes. The detailed IPW search procedure follows Zhang and Wang (2016).

Because each IPW consists of multiple connected road segments, its serviceability at time $t$ is determined by the service levels of its constituent segments. For the $k$th IPW $P_k(i,j)$ between nodes $i$ and $j$, the pathway serviceability is defined as

$$S_k^t(i,j) = \begin{cases} \frac{\sum_{l\in P_k(i,j)} L_l s_l^t}{\sum_{l\in P_k(i,j)} L_l}, & \min_{l\in P_k(i,j)} s_l^t \neq 0, \\ 0, & \min_{l\in P_k(i,j)} s_l^t = 0. \end{cases} \tag{8}$$

where $L_l$ denotes the length of road segment $l$. When all constituent road segments remain passable, $S_k^t(i,j)$ represents their road-length-weighted service level. If any constituent segment becomes impassable, the serviceability of the entire IPW is set to zero. The pathway-level state therefore accounts for both flood-induced degradation in travel efficiency and loss of pathway continuity.

The WIPW-based functionality of road node $i$ at time $t$ is then calculated as

$$NIPW_i^t = \frac{1}{n-1}\sum_{\substack{j=1 \\ j\neq i}}^{n} w_j \left[\sum_{k=1}^{K_{(i,j)}^t} w_k(i,j) S_k^t(i,j)\right] \tag{9}$$

where $n$ is the total number of nodes in the road network; $K_{(i,j)}^t$ is the total number of IPWs between nodes $i$ and $j$ at time $t$; $S_k^t(i,j)$ is the serviceability of the $k$th IPW; $w_k(i,j)$ is the corresponding pathway weight; and $w_j$ is the weight assigned to destination node $j$. Through this aggregation, $NIPW_i^t$ incorporates the service conditions and redundancy of the pathways connecting node $i$ to other destinations, together with the relative importance of those destinations.

The pathway weight $w_k(i,j)$ accounts for pathway length and traffic use and is calculated as

$$w_k(i,j) = 0.5\, L'_{P_k(i,j)} + 0.5\, T'_{P_k(i,j)} \tag{10}$$

where $L'_{P_k(i,j)}$ and $T'_{P_k(i,j)}$ denote the normalized pathway-length and average daily traffic (ADT) terms, respectively. The two components are assigned equal importance in this study, consistent with Wang and Wang (2024), and their detailed normalization procedures follow Zhang and Wang (2016).

The destination-node weight $w_j$ represents the relative importance of node $j$ in emergency-response operations. In the adopted WIPW framework, nodes associated with critical facilities, such as hospitals and fire stations, are distinguished from other major destinations, including residential areas, economic hubs, and key intersections. The same functional-category weighting scheme was applied across all study regions, with nodes serving the same functional role assigned the same weight and no region-specific recalibration performed. For a given G-R2R model, these weights are specified before RNF-label generation according to the intended decision priorities and remain fixed throughout training and inference.

Finally, $NIPW_i^t$ was normalized by the corresponding functionality under normal operating conditions to obtain the node-level RNF state used by G-R2R:

$$q_i^t = \frac{NIPW_i^t}{NIPW_i^0} \tag{11}$$

where $NIPW_i^0$ denotes the normal-state functionality of node $i$. Accordingly, $q_i^t \in [0,1]$, with $q_i^t = 1$ indicating retention of the normal-state functionality and progressively smaller values indicating greater functionality degradation; the corresponding relative functionality loss is $1 - q_i^t$. The baseline $NIPW_i^0$ can be computed from the target region's baseline road-network information using the same WIPW formulation. Therefore, when the corresponding unnormalized NIPW value is required, it can be reconstructed from the predicted relative state as $NIPW_i^t = q_i^t NIPW_i^0$. The normalized node-level states are subsequently assembled into time-varying RNF sequences and used as simulation-derived labels for G-R2R training and evaluation.

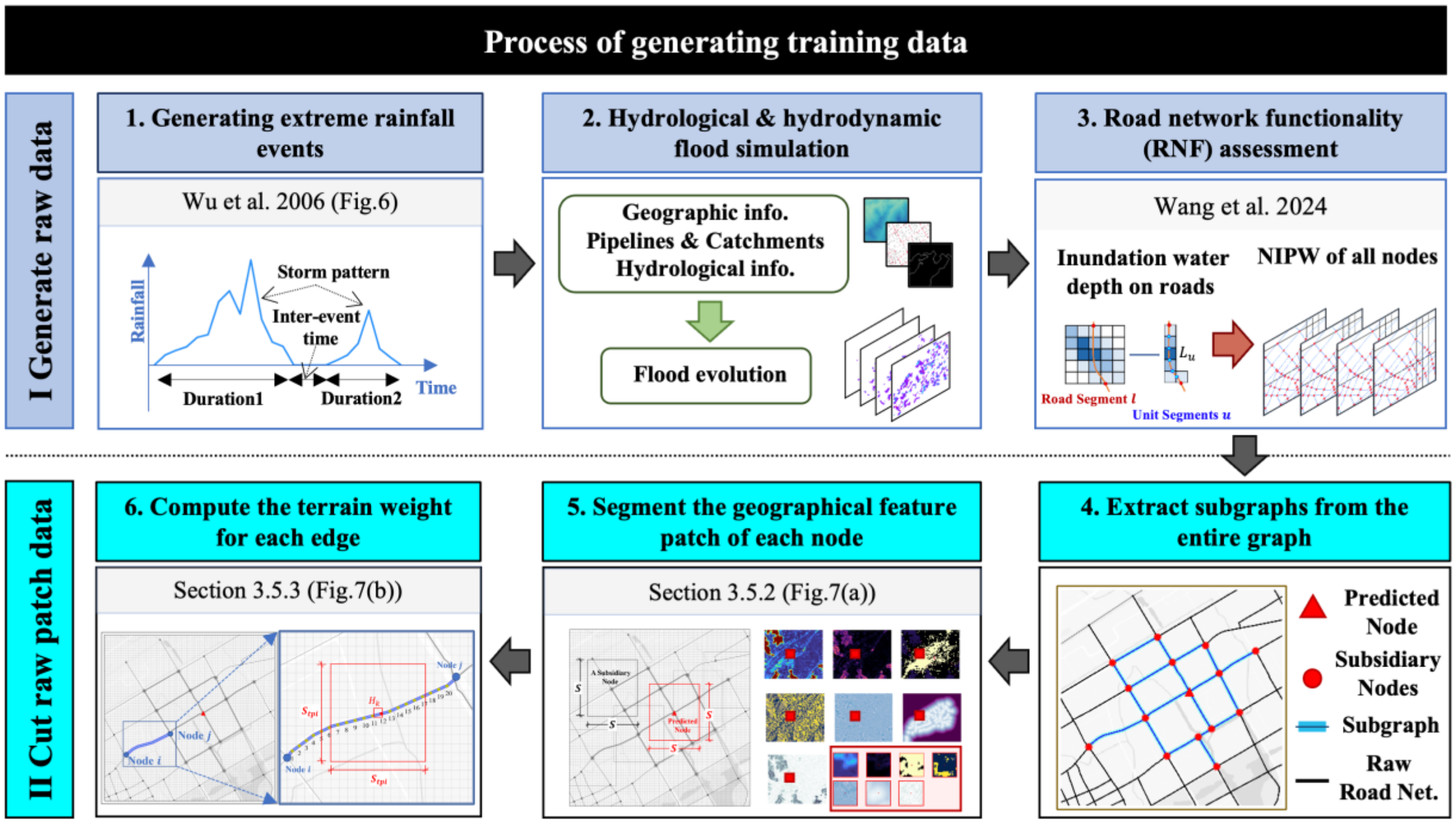

**Fig. 5. Simulation-Derived data generation and graph sample construction workflow for G-R2R.** The upper row generates dynamic RNF labels through stochastic rainfall simulation, flood modeling, and road-network functionality assessment; the lower row transforms the simulation outputs into topology-flexible graph samples.

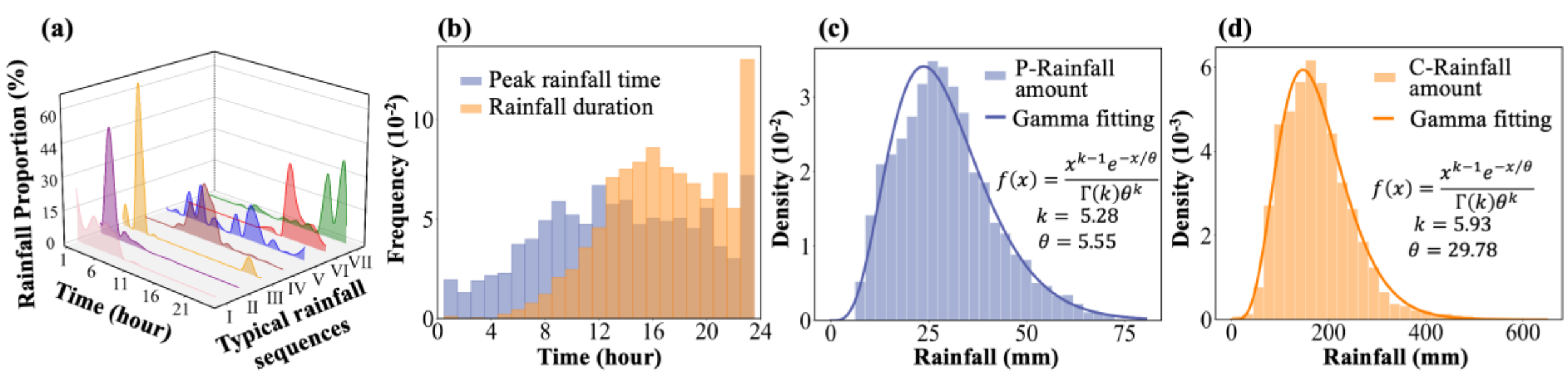

**Fig. 6.** Statistical characteristics of the generated extreme-rainfall events, including representative temporal profiles, peak timing, peak intensity, and total accumulation.

### 3.5. Topology-Flexible Graph Sample Construction

After generating simulation-derived RNF labels, the next step is to transform the raw simulation outputs and static geographic layers into graph samples suitable for G-R2R training. This step implements the topology-flexible graph representation introduced in Section 2. Instead of training on one fixed road-network graph, G-R2R operates on localized subgraphs that can be extracted from road networks with different sizes, densities, and connectivity structures. As shown in the lower row of Fig. 5, the graph sample construction process consists of three steps: ego-centered subgraph sampling, raster-based node feature extraction, and terrain-conditioned adjacency construction.

#### 3.5.1 Ego-Centered Subgraph Sampling

To accommodate road networks with different sizes while maintaining a consistent model input structure, G-R2R represents each prediction target using a fixed-size local ego-centered subgraph. Rather than operating directly on the complete road network, a separate local subgraph is constructed around each target node. This target-centered representation allows road networks with different node counts, densities, and layouts to be transformed into graph samples with a consistent input structure.

For each target node, breadth-first search (BFS) is performed over the original road-network graph, with the target node serving as the root. The search expands outward hop by hop using a first-in-first-out queue until the prescribed number of nodes $N_S$ is reached. Within each BFS layer, nodes are visited in ascending node-index order. After node sampling, all original road-network edges whose endpoints are both contained in the sampled node set are retained to form the local subgraph. This construction preserves the original connectivity relationships among the sampled nodes and produces a connected neighborhood around the target node. To balance model performance and training cost, the subgraph size was selected based on a sensitivity analysis. The results indicate that increasing the subgraph size can further improve prediction performance, while the marginal gain gradually diminishes as the subgraph becomes larger. Accordingly, $N_S = 15$, including the target node itself, is adopted in the main experiments. The corresponding sensitivity results are reported in Section 5.4.

During sample generation, each road node in a source region is successively used as the target node, and an ego-centered subgraph is constructed around it. The target node serves as the prediction target, while the remaining sampled nodes provide the surrounding network context. During inference, the same procedure is applied to all nodes in the target road network. The RNF predictions obtained for individual target nodes are then assembled to represent the functionality state of the complete road network.

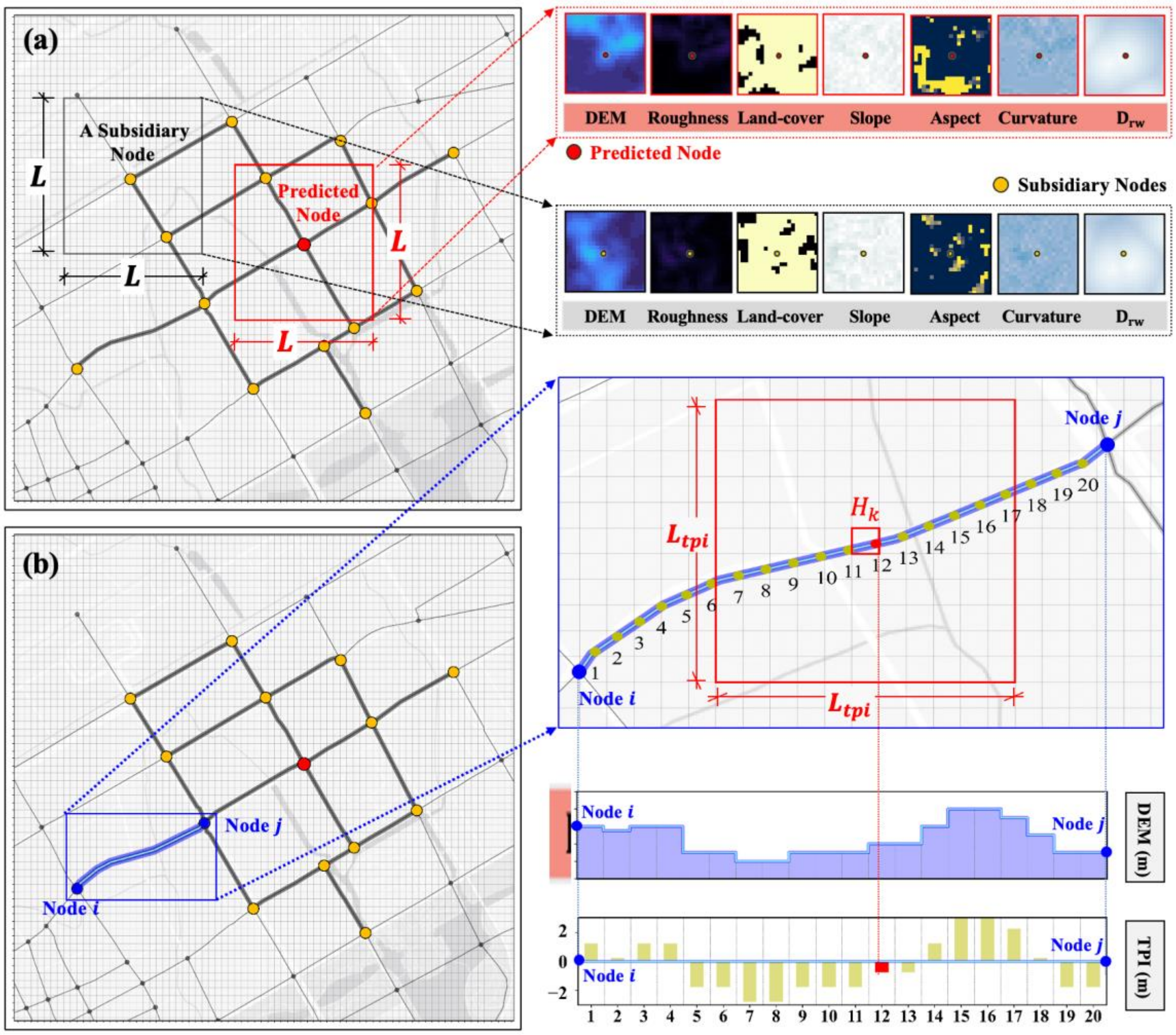

**Fig. 7. Topographic feature extraction and terrain-conditioned adjacency construction for G-R2R graph samples.** Panel (a) illustrates raster-based feature extraction around road nodes; panel (b) illustrates road-segment terrain-position computation using DEM-derived terrain position information.

### 3.5.2 Raster-Based Node Feature Extraction

Each target node and its neighboring nodes are linked to raster-based static feature patches that describe local geographic and hydrological context. As shown in Fig. 7(a), a square window with a side length of 600 m, corresponding to 20 DEM grid cells at 30-m resolution, is extracted around each node. The patch contains the raster-based static attributes introduced in Section 3.3, including terrain, land-cover, and drainage-proximity features.

The 600-m window is chosen to balance contextual coverage and redundancy. A smaller window may miss relevant local terrain variation that shapes ponding and runoff concentration, whereas a substantially larger window may increase overlap among neighboring samples and inflate computational cost. The resulting node-level raster patches provide the static geographic inputs for the dual-level static encoders in G-R2R, allowing the model to distinguish nodes that may experience different flood responses even under similar rainfall forcing.

#### 3.5.3 Terrain-Conditioned Adjacency Construction

In addition to node-level static features, G-R2R incorporates road-segment terrain information into the graph structure through a terrain-conditioned adjacency matrix. Conventional binary adjacency represents road connectivity but does not distinguish the relative terrain positions of connected road segments, which can influence their susceptibility to rainfall-induced disruption. We therefore use the Terrain Position Index (TPI) proposed by Weiss (2001) to condition the edge weights while preserving the underlying road-network topology. Each road segment $(i, j)$ is divided into DEM units, and the TPI for each unit k is calculated as:

$$TPI_k = H_k - \frac{1}{L_{tpi}^2}\sum_{z=1}^{L_{tpi}^2} H_z \tag{12}$$

where $H_k$ is the elevation of unit $k$, and $L_{tpi}$ is the side length of the surrounding reference window, set to 10 DEM cells, or 300 m, in this study. Negative TPI values indicate that a location lies below its surrounding terrain and therefore has greater potential flood susceptibility.

For each road segment $(i, j)$, the median TPI across all intersecting DEM units is used as a robust segment-level descriptor of relative terrain position:

$$T_{ij} = Median\left(TPI_k \middle| k \in K_{(i,j)}\right) \tag{13}$$

where $K_{(i,j)}$ is the set of DEM-aligned units intersecting the road segment $(i, j)$. The median reduces sensitivity to isolated terrain extremes while retaining the representative relative terrain position of the road segment.

To construct a consistent terrain-conditioned edge weighting across regions, the dispersion of road-segment TPI values in the training source regions is characterized using the interquartile range:

$$\Delta_{TPI} = IQR\left(T_{ij} \middle| (i,j) \in \text{E in the training source regions}\right) \tag{14}$$

where $IQR(\cdot)$ denotes the interquartile range. $\Delta_{TPI}$ provides a robust scale for normalizing terrain-position differences and is estimated exclusively from the training source regions. The same value is retained when constructing graph samples for the validation and unseen target regions.

The terrain-conditioned edge weight is then defined as:

$$a_{ij} = \begin{cases} exp\left[-\frac{\max(0,-T_{ij})}{\Delta_{TPI}}\right], & (i,j) \in E \\ 0, & (i,j) \notin E \end{cases} \tag{15}$$

The resulting adjacency matrix is defined as $A = [a_{ij}]$. For road segments with $T_{ij} \geq 0$, the baseline edge weight is retained at $a_{ij} = 1$. For segments with $T_{ij} < 0$, the edge weight decreases monotonically as the representative terrain position becomes lower. This construction therefore preserves the support of the physical road network while reweighting the relative spatial relationships among connected nodes according to road-segment terrain position. Since $0 < a_{ij} \leq 1$ for each existing road segment and $a_{ij} = 0$ otherwise, the resulting adjacency matrix is non-negative. The same terrain-conditioned adjacency matrix $A$ is used in the spectral graph convolution of all ST-Conv blocks in G-R2R, including those in the dynamic rainfall-network encoder and the spatiotemporal decoder described in Sections 4.3 and 4.4. Its contribution to unseen-region RNF forecasting is further evaluated in the ablation analysis by comparison with binary and distance-weighted adjacency constructions.

Together, these steps produce the graph samples used by G-R2R. Each sample contains a target node, its local road-network neighborhood, raster-based static features, terrain-conditioned edge weights, dynamic rainfall inputs, and simulation-derived RNF labels. This construction connects the formulation in Section 2 to the model architecture in Section 4 by providing inputs that are both topology-flexible and region-conditioned, while allowing the same sample structure to be generated from road networks with different sizes and structures without requiring a fixed global graph.

## 4. G-R2R Model Architecture

### 4.1. Architecture Overview

Based on the topology-flexible graph samples constructed in Section 3, G-R2R maps rainfall forcing, recent RNF states, static regional attributes, and terrain-conditioned road-network structure to future node-level RNF sequences. The architecture implements the three modeling requirements identified in Section 2: topology-flexible graph representation, region-conditioned static representation, and dynamic rainfall–network feature learning.

As shown in Fig. 8, G-R2R consists of three modules: a region-conditioned static feature encoder, a dynamic rainfall–network feature encoder, and a spatiotemporal decoder with forecast-horizon control. The static encoder embeds node- and subgraph-level geographic and infrastructure attributes that condition how rainfall forcing translates into local functionality loss. The dynamic encoder learns rainfall-driven RNF evolution over the sampled road-network graph. The decoder then fuses static and dynamic representations to generate future RNF forecasts. Through this design, G-R2R separates time-invariant regional context from time-varying network response, while allowing both to interact during prediction.

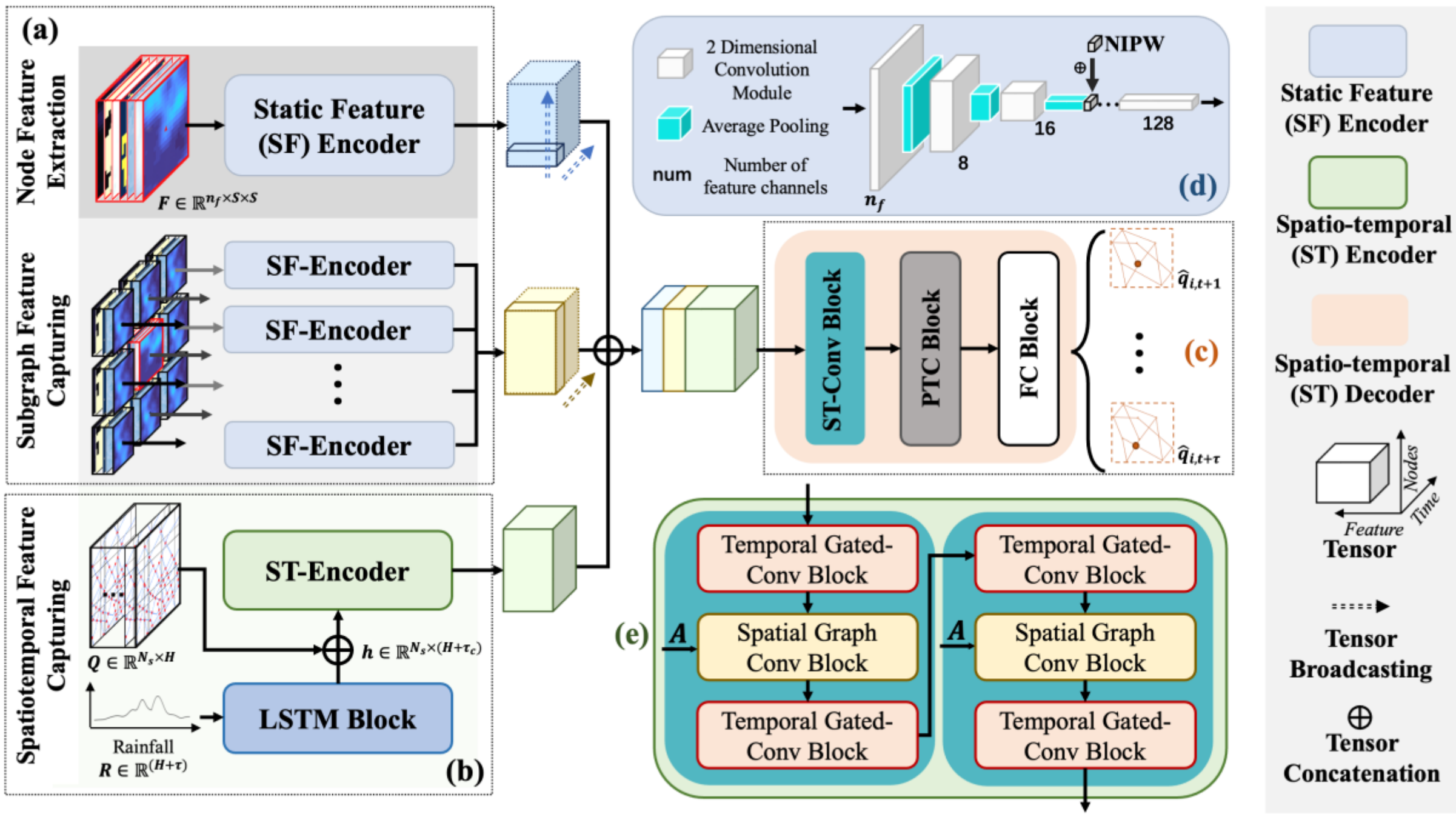

**Fig. 8.** Architecture of the G-R2R Model

### 4.2. Region-Conditioned Static Feature Encoding

Rainfall-induced RNF degradation depends strongly on static regional conditions. Nodes with similar rainfall exposure may experience different functionality loss because of differences in terrain, land-cover, drainage proximity, road elevation, and network redundancy. The static feature encoder therefore provides region-conditioned static representations that account for differences in regional context.

As shown in Fig. 8(a), this module uses a dual-level CNN structure to extract static representations from both the target node and its surrounding subgraph. For each road node, the input is a raster-based static feature tensor $F \in \mathbb{R}^{n_f \times S \times S}$, where $n_f$ denotes the number of static feature layers and $S$ is the spatial patch size. The static layers include the terrain, land-cover, and drainage-proximity attributes introduced in Section 3.3, and the patch size follows the node-centered raster window defined in Section 3.5.2.

Each static feature encoder applies stacked two-dimensional convolution and pooling layers to extract terrain-conditioned spatial features. After the spatial resolution is reduced to $1 \times 1$, the encoded feature vector is concatenated with the normal-state node-level network-performance indicator $NIPW_i^0$, which summarizes the node's redundancy, connectivity, and travel-efficiency role under normal operating conditions. This fusion captures both physical exposure and network importance.

The encoder produces a target-node representation $\boldsymbol{z}_{node} \in \mathbb{R}^{128}$. The same encoder is applied to the nodes within the sampled subgraph, producing a subgraph-level representation $\boldsymbol{Z}_{sub} \in \mathbb{R}^{N_s \times 128}$, where $N_s$ is the number of nodes in the sampled subgraph. In the main experiments, $N_s = 15$, including the target node itself, and the subgraph nodes are sampled using the BFS procedure described in Section 3.5.1. The target-node and subgraph-level representations are then broadcast along the temporal dimension for fusion with dynamic rainfall-RNF features during decoding.

This dual-level design is central to region-conditioned transferability. The node-level representation captures local vulnerability around the prediction target, while the subgraph-level representation captures neighborhood-scale geographic and infrastructure context. Together, they allow G-R2R to distinguish road nodes that may appear similar in topology but respond differently to rainfall because of different static regional conditions.

### 4.3. Dynamic Rainfall–Network Feature Encoding

While static features define regional vulnerability, RNF degradation and recovery unfold dynamically as rainfall evolves. The dynamic rainfall–network feature encoder captures this process by combining rainfall sequence abstraction with spatiotemporal graph learning over the sampled road network.

As shown in Fig. 8(b), this module includes a rainfall LSTM block and a spatiotemporal graph convolutional network.

**Rainfall LSTM encoding**

The rainfall input $\boldsymbol{R} \in \mathbb{R}^{(H+\tau)}$ concatenates H hours of historical rainfall and τ hours of forecast rainfall. The LSTM block compresses this sequence into a latent rainfall representation $\boldsymbol{r} \in \mathbb{R}^{\tau_c}$ , where $\tau_c = 4$ is the temporal compression length. This representation summarizes rainfall temporal information within the input-forecast window, including intensity accumulation, delayed peaks, and rainfall concentration.

To support node-level graph learning, the rainfall representation is broadcast to all nodes in the sampled subgraph and concatenated with the historical RNF sequence $\boldsymbol{Q} \in \mathbb{R}^{(N_s \times H)}$. The resulting input provides each subgraph node with both recent network-state information and rainfall forcing for downstream spatiotemporal encoding.

**ST-GCN encoding**

The ST-GCN encoder learns spatial dependencies in rainfall-driven functionality changes over the sampled road-network graph. It applies stacked ST-Conv blocks, each combining a temporal gated convolution block and a spectral graph convolution block, as shown in Fig. 8(e). The temporal gated convolution captures nonlinear temporal dependencies in RNF evolution, while the spectral graph convolution performs spatial filtering using the normalized graph Laplacian constructed from the terrain-conditioned adjacency matrix $A$ defined in Section 3.5.3.

This design captures two dynamic drivers of RNF evolution: exogenous rainfall forcing and endogenous network interaction. The rainfall representation identifies temporal triggers of disruption, while graph convolution models spatial dependencies in functionality loss and recovery through the terrain-conditioned road-network relationships. The output is a dynamic spatiotemporal representation $\boldsymbol{H}_{dyn}$ over the sampled subgraph.

### 4.4. Spatiotemporal Decoding and Forecast-Horizon Control

The spatiotemporal decoder fuses static regional priors and dynamic rainfall–network representations to generate future RNF forecasts. As shown in Fig. 8(c), it consists of three components: feature fusion through an ST-Conv block, a Prediction Time Control block for temporal alignment, and a fully connected layer for output projection.

**Feature fusion**

The decoder first concatenates the dynamic representation $\boldsymbol{H}_{dyn}$ with the broadcast static representations $\boldsymbol{z}_{node}$ and $\boldsymbol{Z}_{sub}$. The concatenated representation is then processed by an ST-Conv fusion block, whose spectral graph convolution uses the same terrain-conditioned adjacency matrix $A$ defined in Section 3.5.3. This fusion conditions the forecast on both evolving rainfall-RNF dynamics and static regional vulnerability. In other words, the forecast is conditioned jointly on dynamic rainfall–RNF information and static regional context.

**Prediction Time Control (PTC) block**

Because stacked temporal convolutions reduce the temporal dimension, the decoder requires a temporal alignment mechanism to map internal feature representations to the desired forecast horizon. The PTC block performs this function using a one-dimensional convolution over the temporal dimension. Its kernel size $k_p$ is defined as:

$$k_p = H + \tau_c - 6(k_t - 1) - \tau + 1 \tag{16}$$

where $k_t$ is the temporal gated convolution kernel size, $H$ is the historical RNF input length, $\tau_c$ is the rainfall compression length, and $\tau$ is the forecast horizon. The term $6(k_t - 1)$ accounts for the cumulative reduction in temporal length caused by the six temporal gated-convolution operations preceding the PTC block. This design aligns the internal temporal representation with the target output window and decouples the encoder's temporal receptive field from the forecast horizon.

The PTC block enables fixed-horizon RNF prediction while preserving a flexible internal representation of rainfall and network-state evolution. When combined with the iterative prediction strategy evaluated in Section 5, it supports recursive forecasting over longer event periods.

**Output projection**

Finally, a fully connected layer maps the fused representation to the forecasted RNF sequence of target node $i$, $\hat{\boldsymbol{q}}_{i,t+1:t+\tau}$. During inference, predictions are generated for each sampled target-node subgraph and assembled into full-network RNF forecasts.

Together, these modules allow G-R2R to integrate static regional context, dynamic rainfall-RNF information, and graph-structured network interactions within a unified forecasting architecture. The following section evaluates whether these design choices improve component-level performance, cross-region generalization, and temporal reliability under iterative forecasting.

## 5. Experimental Design and Model Evaluation

### 5.1. Experimental Setup, Metrics, and Implementation

Using the multi-region graph dataset constructed in Section 3 and the G-R2R architecture described in Section 4, the experiments evaluate the model from six perspectives: comparative performance against representative forecasting models, contributions of model components and graph representations, sensitivity to subgraph size, cross-region transfer performance, the effects of source-region coverage and training-data volume, and temporal stability under iterative forecasting.

Different experimental protocols are adopted according to the evaluation purpose. For the comparative baseline and component analyses, LX is used as the unseen test region, while the remaining nine communities are used as source regions for model development. For cross-region evaluation, each of the ten communities is held out in turn as the unseen test region, with the remaining nine communities used as source regions. For the source-region coverage analysis, LX is fixed as the unseen test region, and the number of source regions used for training is progressively increased within the remaining nine communities according to the predefined sequence. A complementary training-data-volume analysis retains all nine source regions while varying the proportion of available training samples from 10% to 100% using nested subsets. For temporal stability and iterative forecasting, LX is also used as the unseen test region, with the remaining nine communities serving as source regions. For each experiment, the target region is held out exclusively for testing, while source-region data are partitioned at the sequence level into training and validation sets at an 80:20 ratio prior to sliding-window extraction.

To evaluate whether G-R2R provides advantages over representative prior forecasting architectures, we compare it with six baseline models: STGCN (Yu et al., 2018), AGCRN (Bai et al., 2020), D2STGNN (Shao et al., 2022), DCRNN (Li et al., 2018), Graph WaveNet (Wu et al., 2019), and TimesNet (Wu et al., 2023). These models represent classical spatiotemporal graph convolution, adaptive graph learning, dynamic graph learning, diffusion-based recurrent graph modeling, and non-graph temporal forecasting. In addition to the full **G-R2R** model, variants are defined for subsequent comparative and component analyses: **G-R2R-nN**, which removes the node-level static encoder; **G-R2R-nS**, which removes the subgraph-level static encoder; **G-R2R-nNS**, which removes both static encoders. Two additional variants are used to examine the effect of graph representation. **G-R2R-BA** replaces the terrain-conditioned adjacency with a binary adjacency matrix, whereas **G-R2R-DWA** adopts a Gaussian distance-weighted adjacency, where the weight between connected nodes $i$ and $j$ is defined as $a_{ij}^{DWA} = \exp(-(d_{ij}/\sigma_d)^2)$, with $d_{ij}$ denoting their Euclidean distance and $\sigma_d$ the standard deviation of edge distances estimated from the training source regions. All other model components and training settings are kept unchanged.

Model performance is evaluated using mean squared error (MSE), mean absolute error (MAE), stabilized mean absolute percentage error (MAPE), coefficient of determination $R^2$, residual distributions, and residual exceedance rates. MSE and MAE measure absolute prediction error, while $R^2$ evaluates agreement between predicted RNF and the simulation-derived RNF reference. Because RNF values can approach zero under severe disruption, a stabilized MAPE is used to avoid numerical inflation caused by division by near-zero targets:

$$MAPE = \frac{1}{n}\sum_{i=1}^{n}\left|\frac{Q_i - \hat{Q}_i}{max\{|Q_i|, \epsilon\}}\right| \times 100\% \tag{17}$$

where $Q_i$ and $\widehat{Q}_i$ denote the reference and predicted RNF values, respectively, and $\epsilon = 0.01$ is the lower bound. Residuals are defined as $e_i = Q_i - \widehat{Q}_i$. Residual exceedance rates are reported to quantify the frequency of large prediction errors under multiple thresholds, which is important for flood-impact forecasting because rare but large errors can lead to misleading emergency assessments.

All forecasting experiments use a historical input window of $H = 12$ h and a forecast horizon of $\tau$=3 h, consistent with the generalized formulation in Eqs. (5)–(7). For each 24-h rainfall–RNF sequence, a 15-h sliding window is used to generate supervised samples. The first 12 h provide the historical rainfall and RNF inputs, while the following 3 h provide the forecast target. The window advances in 3-h increments, producing multiple samples from each simulated rainfall event. Unless otherwise specified, G-R2R uses a subgraph size of $N_s = 15$, including the target node itself; the sensitivity to $N_s$ is examined in Section 5.4. The rainfall LSTM contains two layers with 32 hidden units and blocks with temporal convolution kernel size $k_t = 3$, and the PTC block uses $k_p = 2$ to align the encoded temporal representation with the 3-h forecast horizon. Training is performed using the RMSprop optimizer to minimize MSE loss. The batch size is 1000, the initial learning rate is $10^{-3}$, and the learning rate decays by a factor of 0.7 every five epochs. A dropout rate of 0.2 is applied to reduce overfitting, and early stopping is triggered when validation error does not improve for ten consecutive epochs. All experiments are implemented in PyTorch on a workstation with an Intel Xeon Silver 4314 CPU and an NVIDIA GeForce RTX 4090 GPU.

### 5.2. Comparative Baseline Evaluation

The comparative evaluation follows the fixed unseen-region benchmark protocol described in Section 5.1, with LX used as the unseen test region. Six representative forecasting models—STGCN, AGCRN, $D^2$STGNN, DCRNN, Graph WaveNet, and TimesNet—are compared with the full G-R2R model. G-R2R-nNS, which removes both the node-level and subgraph-level static encoders while retaining the remaining G-R2R structure, is also reported to help distinguish the contribution of the dual-level static encoding from that of the remaining forecasting architecture. Table 3 summarizes the prediction performance of all models.

**Table 3.** Comparative performance of G-R2R and baseline models under the fixed unseen-region benchmark in LX.

| Model | MSE | MAE | MAPE | $R^2$ |
|---|---|---|---|---|
| **STGCN** | 0.0049 | 0.0436 | 5.36% | 0.719 |
| **AGCRN** | 0.0038 | 0.0281 | 4.31% | 0.779 |
| **$D^2$STGNN** | 0.0035 | 0.0333 | 4.07% | 0.797 |
| **DCRNN** | 0.0056 | 0.0320 | 5.25% | 0.671 |
| **Graph WaveNet** | 0.0041 | 0.0335 | 4.37% | 0.761 |
| **TimesNet** | 0.0085 | 0.0553 | 7.32% | 0.547 |
| **G-R2R-nNS** | 0.0047 | 0.0322 | 4.36% | 0.740 |
| **G-R2R** | 0.0026 | 0.0197 | 2.33% | 0.855 |

As shown in Table 3, the full G-R2R model achieves the best overall performance across all four evaluation metrics, with an MSE of 0.0026, MAE of 0.0197, MAPE of 2.33%, and $R^2$ of 0.855. Among the baseline models, $D^2$STGNN achieves the lowest MSE and MAPE and the highest $R^2$, whereas AGCRN yields the lowest MAE. Relative to the strongest baseline under each metric, G-R2R reduces MSE by 25.7% compared with $D^2$STGNN, reduces MAE by 29.9% compared with AGCRN, and reduces MAPE by 42.8% compared with $D^2$STGNN. Its $R^2$ is also 0.058 higher than that of $D^2$STGNN. These results indicate that the complete G-R2R framework provides a clear performance advantage under the fixed unseen-region benchmark.

Among the baseline models, $D^2$STGNN and AGCRN exhibit the strongest overall performance. Their comparatively strong results are consistent with the importance of flexibly representing spatial dependencies in rainfall-induced RNF forecasting. AGCRN adaptively learns graph relationships, whereas $D^2$STGNN models dynamically varying spatial interactions and decoupled spatiotemporal representations. These capabilities are well suited to a forecasting task in which the functionality of individual road nodes evolves jointly with the surrounding network rather than as independent temporal sequences. By contrast, TimesNet does not explicitly represent road-network connectivity, and its comparatively larger prediction errors are consistent with the importance of graph-structured spatial dependencies in node-level RNF forecasting.

The comparison with G-R2R-nNS provides additional insight into the source of the performance improvement. Without the two static encoders, G-R2R-nNS achieves an MSE of 0.0047, MAE of 0.0322, MAPE of 4.36%, and $R^2$ of 0.740. Its performance remains comparable to several graph-based forecasting models, but it does not consistently outperform the stronger AGCRN and $D^2$STGNN baselines. When the dual-level static encoding is restored, MAPE decreases from 4.36% to 2.33%, while $R^2$ increases from 0.740 to 0.855. This comparison indicates that the improvement of the full G-R2R cannot be attributed solely to the shared spatiotemporal forecasting structure; the region-conditioned static representation provides substantial complementary information for predicting rainfall-induced RNF responses across regional road-network contexts.

Taken together, these results are consistent with the motivation of G-R2R: rainfall forcing and recent RNF states alone do not fully characterize how a regional road network responds to extreme rainfall. Static geographic and network context provides additional information for distinguishing different RNF responses under otherwise similar dynamic forcing. The specific contributions of the dual-level static encoders and the terrain-conditioned graph representation are examined separately in the ablation analysis in Section 5.3.

### 5.3. Ablation Analysis of Model Components

Fig. 9(a) shows the MSE and MAE of the full G-R2R model and the five variants defined in Section 5.1. The full G-R2R achieves the lowest prediction errors among all configurations. Among the adjacency variants, G-R2R-DWA performs numerically slightly better than G-R2R-BA across all four metrics, but neither reaches the prediction accuracy obtained with the terrain-conditioned adjacency. The performance advantage observed in Fig. 9(a) is also reflected in MAPE and $R^2$. G-R2R achieves a MAPE of 2.33% and an $R^2$ of 0.855, compared with 4.31% and 0.708 for G-R2R-BA and 4.26% and 0.716 for G-R2R-DWA, respectively. Compared with either conventional adjacency variant, the terrain-conditioned adjacency reduces MSE by approximately 50% and MAPE by approximately 45%. These results indicate that incorporating relative topographic context into the weighting of existing road-network connections provides a more effective graph representation for RNF forecasting than either binary adjacency or conventional distance-weighted adjacency under the present benchmark.

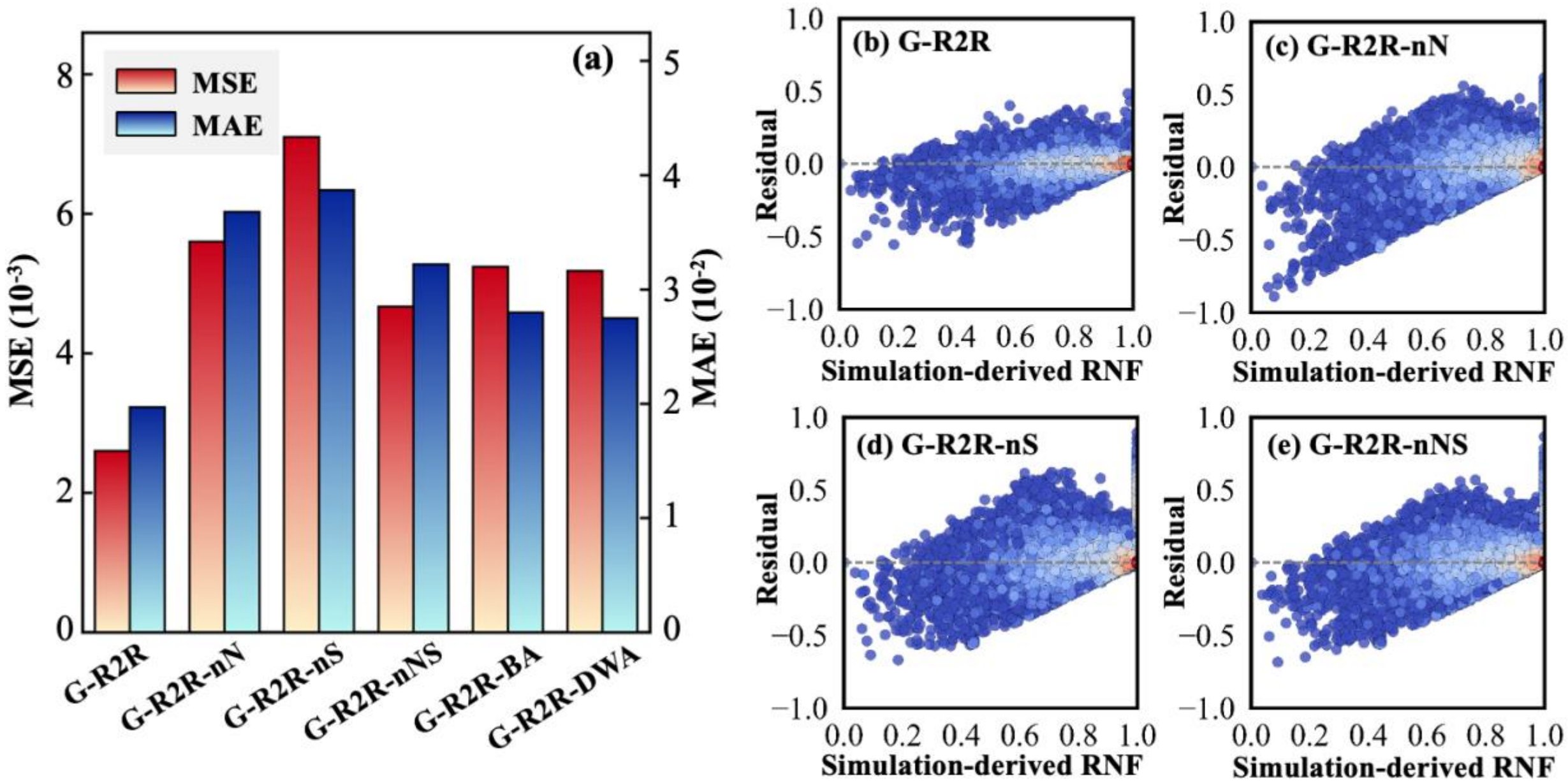

**(f) Residual exceedance rates at multiple absolute-residual thresholds for 4 models**

| Model | $\|Res\|>0.1$ | $\|Res\|>0.2$ | $\|Res\|>0.3$ | $\|Res\|>0.4$ | $\|Res\|>0.5$ |
|---|---|---|---|---|---|
| G-R2R | 2.91% | 0.45% | 0.10% | 0.03% | 0.01% |
| G-R2R-nN | 11.01% | 3.49% | 1.13% | 0.33% | 0.15% |
| G-R2R-nS | 12.38% | 4.16% | 1.89% | 0.91% | 0.39% |
| G-R2R-nNS | 9.70% | 2.97% | 0.98% | 0.37% | 0.15% |

**Fig. 9. Ablation analysis of G-R2R components.** (a) MAE and MSE for the full G-R2R model and five variants; (b)-(e) residual distributions for the full model and the three static-encoding variants; (f) residual exceedance rates for the same four models under different absolute-error thresholds.

The static-encoding variants further show the contribution of the dual-level regional representation. Removing both static encoders increases MAPE from 2.33% to 4.36% and decreases $R^2$ from 0.855 to 0.740. The two partially ablated variants, G-R2R-nN and G-R2R-nS, also perform worse than the full model. Interestingly, G-R2R-nNS performs better than either partially ablated configuration. This non-monotonic ablation pattern indicates that the benefit of the static representation is not obtained by retaining only one of the two encoding levels. Rather, the full dual-level design provides complementary target-node and neighborhood-scale contextual information that is most effective when used jointly.

The residual distributions in Fig. 9(b)-(e) further show that dual-level static encoding improves prediction performance. The full G-R2R model produces the most concentrated residual distribution and substantially fewer large deviations. By contrast, the partially ablated variants exhibit wider residual spreads, especially at low RNF values corresponding to severe network degradation. This pattern suggests that static geographic and network context becomes especially informative when network functionality is strongly disrupted and local responses differ across regional conditions.

Fig. 9(f) quantifies this improvement using residual exceedance rates under multiple absolute-error thresholds. The full G-R2R model consistently yields the lowest exceedance rates, decreasing from 2.91% at the 0.1 threshold to 0.01% at the 0.5 threshold. The ablated variants show substantially heavier error tails; for example, G-R2R-nS has an exceedance rate of 12.38% at the 0.1 threshold and 0.39% at the 0.5 threshold. This indicates that removing one or both static encoders not only increases average error but also increases the frequency of large prediction errors. Overall, the ablation results support the contributions of both the terrain-conditioned graph representation and the dual-level static encoding to G-R2R performance.

### 5.4. Sensitivity to Subgraph Size

To examine the sensitivity of G-R2R to the subgraph size $N_s$, we evaluate four settings, $N_s$=5, 10, 15, and 20, under the fixed unseen-region benchmark described in Section 5.1. Table 4 summarizes the corresponding prediction performance. The results show a clear improvement as $N_s$ increases from 5 to 15. In particular, MAPE decreases from 6.34% to 2.33%, while $R^2$ increases from 0.545 to 0.855. This trend suggests that a broader local neighborhood provides useful contextual information for target-node RNF prediction within the tested range.

When $N_s$ is further increased from 15 to 20, the overall prediction performance remains at a similar level, with no consistent additional gain across the evaluation metrics. This suggests that enlarging the sampled neighborhood beyond 15 nodes provides limited additional predictive benefit under the present benchmark. Taken together, these results support $N_s = 15$ as a suitable setting for the main experiments, as it provides the best overall predictive performance among the tested subgraph sizes, while increasing $N_s$ to 20 yields no consistent additional improvement.

**Table 4.** Sensitivity of G-R2R performance to subgraph size $N_s$.

| $N_s$ | MSE | MAE | MAPE | $R^2$ |
|---|---|---|---|---|
| 5 | 0.0082 | 0.0458 | 6.34% | 0.545 |
| 10 | 0.0052 | 0.0353 | 4.85% | 0.711 |
| 15 | **0.0026** | **0.0197** | **2.33%** | **0.855** |
| 20 | **0.0026** | 0.0266 | 2.36% | 0.854 |

### 5.5. Cross-Region Generalization

To evaluate whether G-R2R can transfer across spatially heterogeneous communities, we conducted leave-one-area-out cross-validation across the ten flood-prone regions. In each iteration, one community was held out as an unseen test region, while the remaining communities were used as source regions for model training and validation. No samples from the held-out region were used during training. This protocol directly tests the model's ability to forecast RNF in previously unseen road-network domains.

Fig. 10(a)-(b) summarizes the model's predictive performance across the ten held-out regions. Overall, G-R2R achieves consistently low errors across all unseen regions, although moderate variability is observed among communities. This variability is consistent with differences in terrain, drainage context, road-network topology, and baseline network performance among the study regions, which together shape the difficulty of predicting rainfall-induced RNF degradation.

The stabilized MAPE values range from 2.30% to 3.54%, indicating strong predictive accuracy across all held-out regions. The $R^2$ values are also consistently high, suggesting close agreement between G-R2R predictions and the simulation-derived RNF references. The density-based scatter plots for representative regions, including LX, LG, OH, and PY, further show that predicted RNF values align closely with the 1:1 reference line, with limited systematic bias.

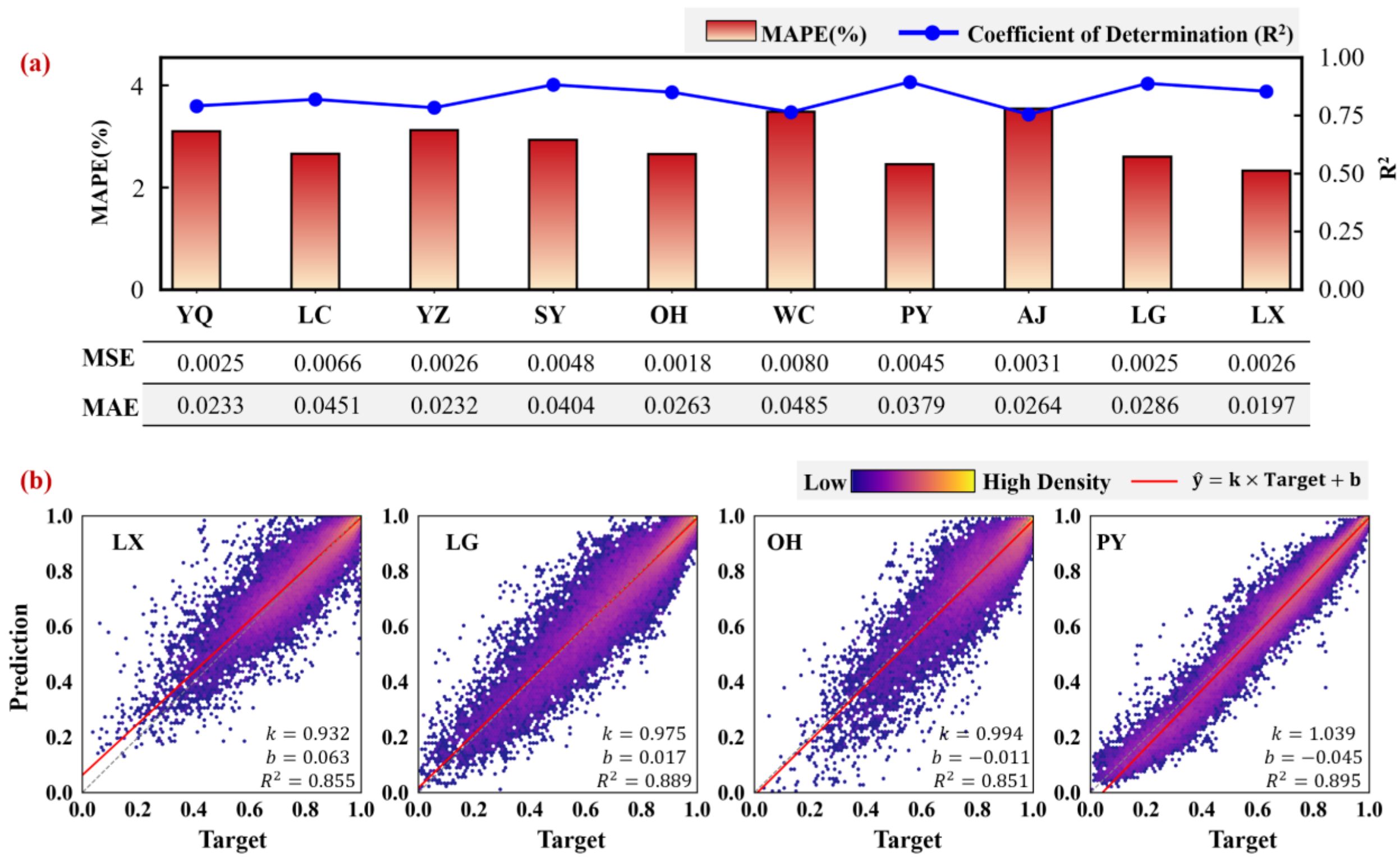

**Fig. 10. Cross-region generalization performance of G-R2R.** (a) MAPE, MSE, MAE, and $R^2$ across the ten held-out regions. (b) Density scatter plots comparing predicted and target RNF values for four representative regions, with fitted regression lines shown in red.

These results demonstrate that G-R2R achieves cross-region generalization across the regional road-network domains represented in this study, rather than merely fitting region-specific rainfall–RNF patterns. The model remains accurate across communities with different static attributes and network structures, indicating that the learned predictive relationships can be transferred to previously unseen regional networks without target-region retraining.

### 5.6. Effects of Source-Region Coverage and Training-Data Volume

To examine how source training information affects unseen-region prediction, we conduct two complementary experiments with LX fixed as the unseen test region. In the first experiment, the source training set is progressively expanded from one to nine regions following the sequence YQ, LC, YZ, SY, OH, WC, PY, AJ, and LG. In the second experiment, all nine source regions are retained, while the available training samples are progressively increased from 10% to 100% using nested subsets.

As shown in Table 5(a), prediction performance generally improves as more source regions are included in the training set. From one to nine source regions, MAPE decreases from 10.99% to 2.33%, while $R^2$ increases from 0.382 to 0.855. The improvement is not strictly monotonic, and temporary fluctuations occur as additional regions are introduced. Nevertheless, the overall tendency is toward lower prediction errors and higher $R^2$ with broader source-region coverage. This suggests that training across a broader set of source-region conditions provides useful information for predicting RNF in the unseen LX region.

A similar overall trend is observed in Table 5(b) when the source-region pool is kept fixed and only the amount of training data is varied. As the training-data proportion increases from 10% to 100%, MAPE decreases from 10.19% to 2.33%, while $R^2$ increases from 0.154 to 0.855. Although small fluctuations remain at several intermediate proportions, prediction accuracy improves substantially as more training samples become available. The nested-subset experiment therefore shows that increasing training-data volume within the same nine-region source pool can substantially improve unseen-region prediction performance.

Together, the two experiments show an overall improvement in unseen-region prediction under the predefined expansion of the source-region pool and with increasing training-data volume. The first experiment reflects the combined influence of source-region composition and increasing training information under the predefined sequence, whereas the second more directly examines the role of training-data volume under a fixed source-region pool.

Within the scope of the present study, these results indicate that unseen-region prediction generally improves with broader source-region coverage and increasing training-data volume. Extending the source-region pool to include a wider range of geographic and hydroclimatic conditions may provide a pathway toward broader transferability, although this remains to be evaluated under more distinct urban and climatic settings.

**Table 5.** Effects of source-region coverage and training-data volume on G-R2R performance in the unseen LX region.

**(a) Source-region coverage**

| No. of source regions | Newly added region | MSE | MAE | MAPE | $R^2$ |
|---|---|---|---|---|---|
| 1 | YQ | 0.0108 | 0.0936 | 10.99% | 0.382 |
| 2 | LC | 0.0079 | 0.0487 | 6.54% | 0.531 |
| 3 | YZ | 0.0058 | 0.0375 | 5.16% | 0.691 |
| 4 | SY | 0.0054 | 0.0361 | 5.06% | 0.695 |
| 5 | OH | 0.0049 | 0.0341 | 4.81% | 0.723 |
| 6 | WC | 0.0060 | 0.0388 | 5.24% | 0.671 |
| 7 | PY | 0.00369 | 0.0287 | 3.26% | 0.794 |
| 8 | AJ | 0.0051 | 0.0353 | 4.93% | 0.701 |
| 9 | LG | **0.0026** | **0.0197** | **2.33%** | **0.855** |

**(b) Training-data volume**

| Training-data proportion | MSE | MAE | MAPE | $R^2$ |
|---|---|---|---|---|
| 10% | 0.0152 | 0.0654 | 10.19% | 0.154 |
| 20% | 0.0115 | 0.0659 | 8.73% | 0.356 |
| 30% | 0.0097 | 0.0498 | 8.02% | 0.458 |
| 40% | 0.0066 | 0.0439 | 5.74% | 0.631 |
| 50% | 0.0059 | 0.0413 | 5.53% | 0.693 |
| 60% | 0.0048 | 0.0364 | 4.97% | 0.708 |
| 70% | 0.0043 | 0.0365 | 5.12% | 0.761 |
| 80% | 0.0030 | 0.0269 | 3.71% | 0.833 |
| 90% | 0.0029 | 0.0281 | 3.84% | 0.840 |
| 100% | **0.0026** | **0.0197** | **2.33%** | **0.855** |

### 5.7. Temporal Stability and Iterative Forecasting

To evaluate temporal reliability under evolving rainfall conditions, we analyze lead-time-dependent prediction errors using LX as the unseen test region. This experiment examines how prediction errors evolve with forecast lead time and during iterative rolling forecasts over an extended prediction period.

Fig. 11(a) shows the distribution of node-level relative errors over the LX test set. Most predictions remain close to the simulation-derived RNF reference, with the central 95% of relative errors falling between -6.7% and 6.1%. The overall MAPE is 2.33%, indicating good overall prediction accuracy over the 3-h forecast horizon. Large relative errors are infrequent and mainly occur in low-RNF regimes, where small absolute deviations can be amplified by near-zero reference values.

Because G-R2R produces a 3-h forecast sequence, we further examine how errors vary within the forecast window. As shown in Fig. 11(b), error dispersion increases slightly with lead time, but the increase remains moderate across the 3-h forecast window. This indicates that the model maintains relatively stable predictive performance across the three forecast steps.

To support longer-duration emergency operations, we also evaluate an iterative rolling-forecast strategy. As illustrated in Fig. 11(c), the model uses its previous RNF prediction as the updated network-state input for the next prediction cycle, while rainfall forcing is provided for the corresponding forecast window. Fig. 11(d) shows the relative-error distributions over 12 consecutive hourly predictions generated through this rollout process. Although error dispersion increases gradually, the increase remains moderate over the 12-h rollout, indicating that G-R2R maintains temporal stability during iterative forecasting.

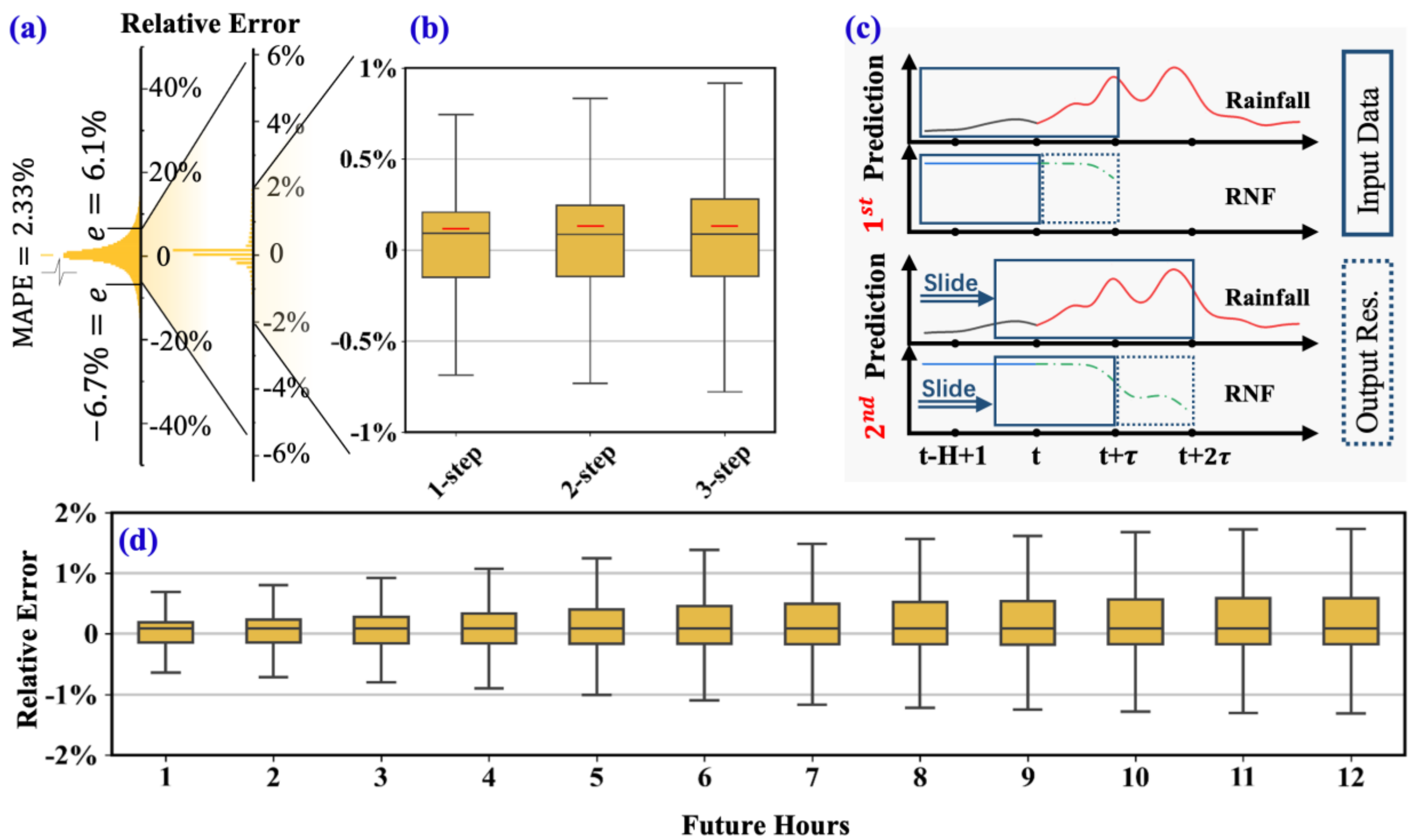

**Fig. 11. Temporal reliability and rolling-forecast performance of G-R2R**: (a) Distribution of node-level relative errors in LX. (b) Relative-error distributions across forecast lead times within the 3-h prediction window. (c) Iterative rolling-forecast strategy, where previous RNF predictions are used as updated network-state inputs. (d) Relative-error distributions over 12 consecutive hourly predictions. The results indicate stable short-horizon accuracy and bounded error growth during iterative forecasting.

These results demonstrate stable short-horizon RNF prediction and support the feasibility of iterative forecasting over longer forecast periods. Such temporal stability is important for applications requiring continuously updated estimates of road-network degradation and recovery as rainfall conditions evolve.

## 6. HISTORICAL-EVENT FORECAST REPLAY IN UNSEEN REGIONS: Typhoon Fitow

### 6.1. Event Setup and Forecasting Protocol

Using the observed rainfall process of Typhoon Fitow, we conduct a historical-event forecast replay to further examine G-R2R under an extreme-rainfall event with realistic temporal forcing. In contrast to the metric-based evaluations in Section 5, this section further examines how the predicted RNF evolves over time and space under a historical event, thereby providing a more direct view of the model outputs in previously unseen regional road networks. RNF predictions are evaluated against the simulation-derived reference generated by the regional flood-simulation and network-functionality assessment workflow described in Section 3.

Fig. 12 shows the track of Typhoon Fitow relative to the study region. Fitow developed into a severe tropical storm over the northwestern Pacific on October 1, made landfall approximately 40 km south of Lingxi at 18:15 UTC on October 6, and dissipated by 06:00 UTC on October 7. The storm produced intense rainfall over multiple Wenzhou communities, providing a representative historical event for examining rainfall-induced RNF degradation and recovery.

G-R2R is applied to four target regions—LX, LG, PY, and OH—without using training samples from the corresponding target region. Forecasting is initialized at 02:00 UTC on October 6. The initial historical RNF state is set to its normal pre-impact condition. During the rollout, previous RNF predictions are used as updated network-state inputs for subsequent prediction cycles. Rainfall forcing is provided by observed station rainfall records from the Zhejiang Meteorological Bureau. In prospective forecasting, the rainfall input can instead be supplied by forecast or nowcast products.

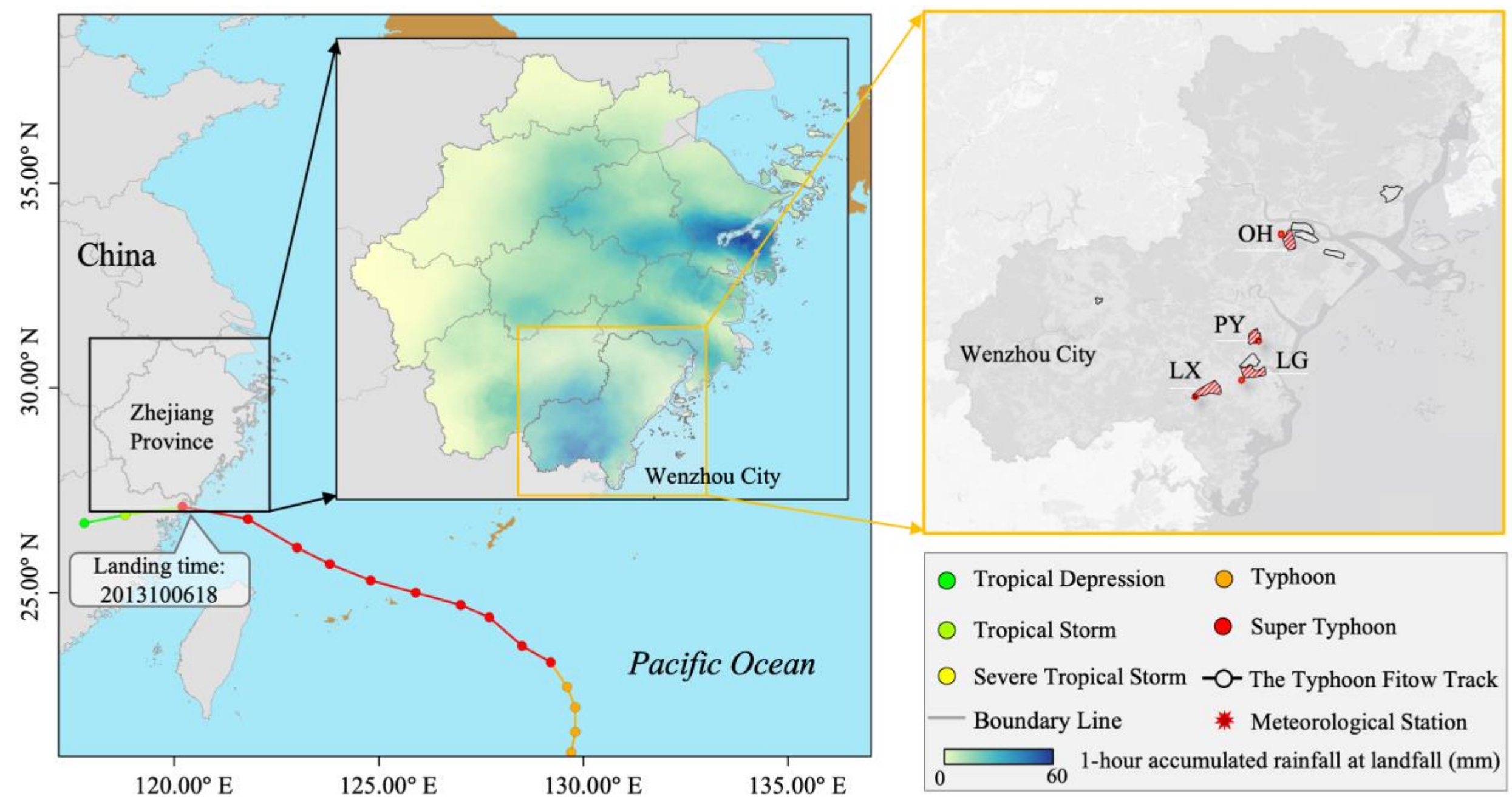

**Fig. 12. Track of Typhoon Fitow and locations of the replay communities.** The map shows the intensity-classified track of Typhoon Fitow, the 1-h accumulated rainfall at landfall, and the locations of the four replay regions in Wenzhou: LX, LG, PY, and OH.

From a transportation-forecasting perspective, this protocol examines whether a model trained on other source regions can capture when road-network functionality begins to degrade, where low-functionality hotspots emerge and evolve, and how network recovery unfolds over time in previously unseen road networks. By examining these predictions through both temporal trajectories and node-level spatial patterns, the historical-event replay provides a direct view of how the model outputs evolve during an extreme-rainfall event beyond metric-based performance evaluation, while requiring neither target-region retraining nor repeated execution of the full flood-simulation and network-functionality workflow.

### 6.2. Temporal Evolution of Region-level RNF

Fig. 13 compares the G-R2R-predicted RNF trajectories with the simulation-derived RNF references for LX, LG, OH, and PY during Typhoon Fitow. The region-level RNF at each time step is calculated as the mean of the node-level RNF values over all road nodes in the corresponding region. Across the four unseen regions, G-R2R reproduces the major phases of RNF evolution, including pre-impact stability, rapid degradation, peak disruption, and post-peak recovery. The close correspondence in these temporal trajectories shows that the model captures not only the severity of functionality loss but also the timing of degradation and recovery during the historical rainfall event. Such temporal information is particularly relevant to emergency assessment because response planning depends not only on the expected severity of network degradation, but also on when deterioration begins, how quickly it develops, and when recovery is likely to occur.

The forecasts also reveal distinct temporal responses among the four regions. LX and LG experience earlier and sharper RNF declines, OH shows delayed degradation followed by a faster rebound, and PY exhibits a more prolonged low-functionality period. These differences are consistent with variations in rainfall forcing and regional geographic, hydrological, and network context. The ability of G-R2R to reproduce these distinct temporal patterns across unseen regions is consistent with its region-conditioned forecasting design, in which dynamic rainfall forcing is interpreted together with static regional and network information.

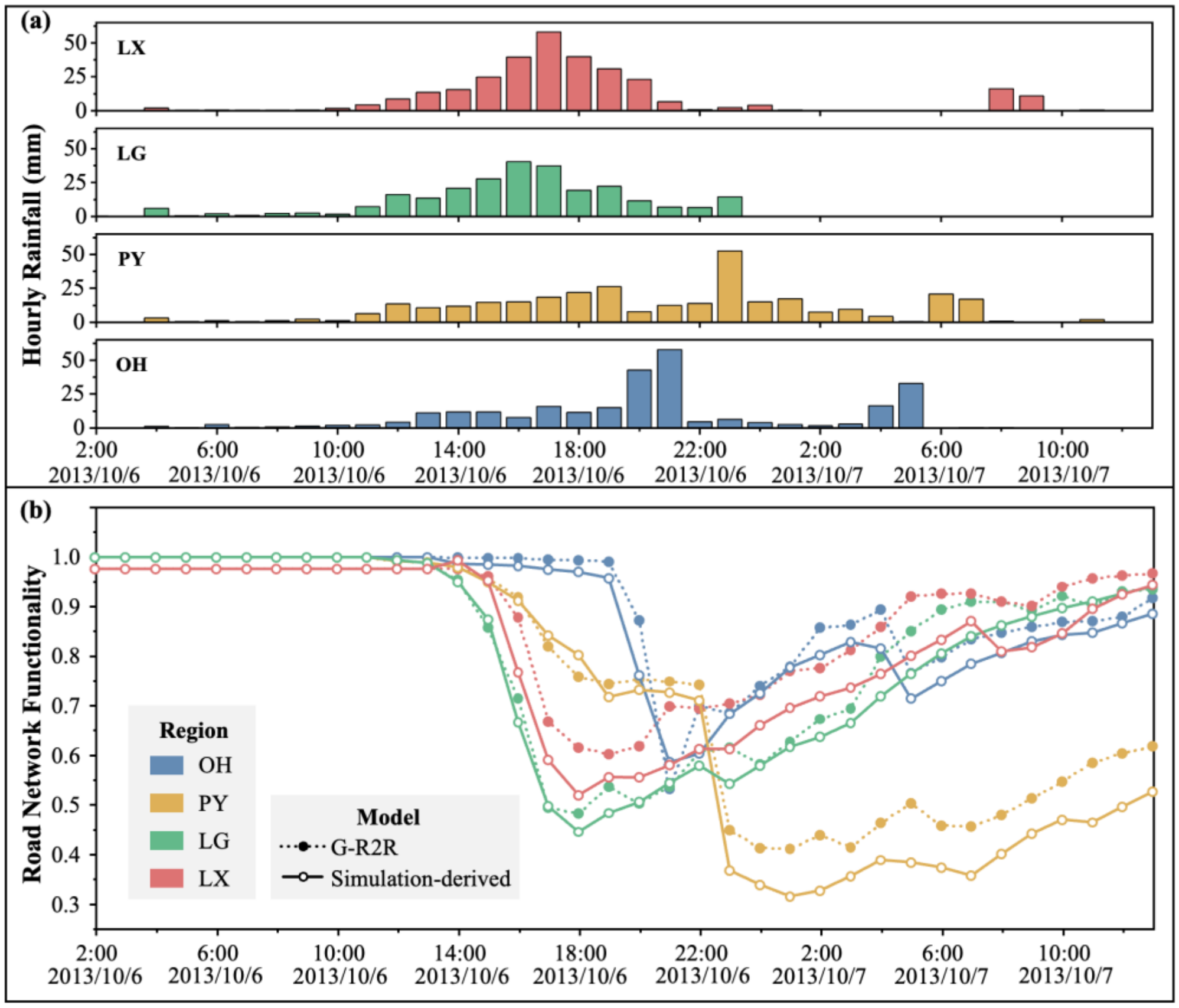

**Fig. 13. Rainfall forcing and region-level RNF trajectories during Typhoon Fitow.** (a) Observed rainfall forcing used in the historical-event forecast replay. (b) Comparison between G-R2R-predicted and simulation-derived reference RNF trajectories for LX, LG, OH, and PY.

### 6.3. Node-Level Hotspot Evolution

Region-level RNF trajectories summarize the overall degradation and recovery of each road network, but emergency response also requires spatially explicit information on where functionality losses concentrate. Fig. 14 extends this evaluation from temporal trajectories to node-level spatial comparison across four representative regions—OH, PY, LG, and LX. For each community, G-R2R predictions are compared with the simulation-derived reference at four representative timestamps, which collectively capture different stages of rainfall-induced RNF loss and recovery during Typhoon Fitow.

Across the four communities, G-R2R reproduces the dominant spatial patterns of RNF degradation shown by the simulation-derived reference. At 14:00 UTC on October 6, most road nodes remain highly functional, with only scattered low-RNF points. By 18:00 UTC, low-functionality nodes begin to expand in several communities, especially in LG and LX, where clustered degradation becomes visible. At 22:00 UTC, the disruption patterns become more pronounced: OH, LG, and LX show clearer spatial aggregation of low-RNF nodes, while PY remains less severely affected. By 02:00 UTC on October 7, inter-community differences become more evident. PY exhibits a delayed and more extensive degradation pattern, whereas OH, LG, and LX show varying degrees of residual disruption and early recovery. Importantly, the G-R2R maps remain spatially consistent with the simulation-derived reference across these different timing, severity, and network-geometry conditions.

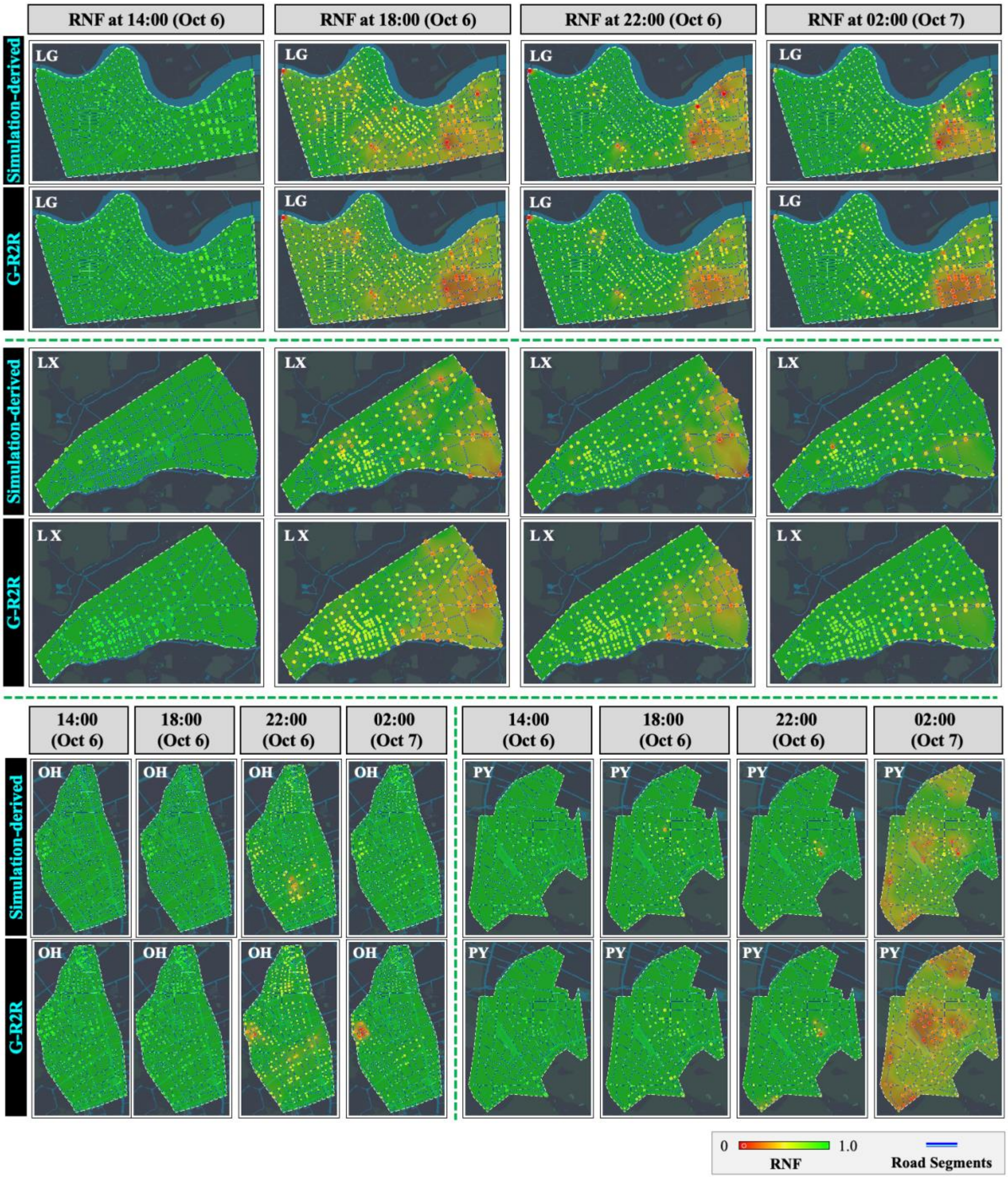

**Fig. 14. Node-level RNF evolution in LG, LX, OH and PY during Typhoon Fitow**. Comparison between simulation-derived reference RNF maps and G-R2R-predicted RNF maps at four representative timestamps spanning rainfall-induced degradation and recovery: 14:00 UTC, 18:00 UTC, and 22:00 UTC on October 6, and 02:00 UTC on October 7.

These multi-region snapshots show that G-R2R captures not only the overall temporal evolution of regional RNF, but also the node-level spatial organization of functionality degradation. The forecasts reveal where low-functionality clusters emerge, how they expand or persist, and where recovery begins as the rainfall event evolves. This spatial information is relevant to emergency assessment because RNF reflects changes in transportation-network functionality associated with connectivity, accessibility, and network redundancy, rather than local flood exposure alone. It therefore provides a network-oriented view of where the functional consequences of rainfall-induced flooding are concentrated.

The node-level RNF forecasts can consequently provide decision-relevant information for network-level functional risk assessment and recovery planning. Areas experiencing substantial RNF degradation indicate locations where network accessibility and redundancy have deteriorated markedly, and can therefore serve as screening information for closer inspection, pre-positioning of emergency repair or support resources, and prioritization during network recovery. Because RNF is normalized relative to each node's normal-state functionality, these spatial patterns characterize the relative degree of functionality degradation rather than absolute network importance. For cross-location prioritization, the predicted RNF should therefore be interpreted together with normal-state network-performance information, which distinguishes the baseline functional roles of different nodes. In this way, RNF forecasts complement hazard-oriented inundation information by translating rainfall-induced flood impacts into transportation-network functional consequences, thereby providing spatially explicit information for emergency preparedness, network-function assessment, and recovery planning during evolving rainfall events.

### 6.4. Peak-Stage Error Characterization

Fig. 15 evaluates G-R2R prediction errors at the peak RNF-loss stage of each target region: 21:00 UTC on October 6 for OH, 18:00 UTC on October 6 for LX and LG, and 01:00 UTC on October 7 for PY. These time points correspond to the strongest functionality degradation observed in each region during the historical-event replay. Fig. 15(a) shows the node-level spatial distribution of relative errors, while Fig. 15(b) relates relative error to the corresponding simulation-derived reference RNF level.

Most road nodes exhibit relatively low prediction errors, while larger deviations are confined to limited localized areas. In OH, higher errors are mainly concentrated around the central–southern degradation hotspot. In LX, the largest errors occur near the eastern boundary, where severe functionality loss is observed. LG shows localized discrepancies in the more strongly affected eastern portion of the network, whereas PY contains only a small number of elevated-error nodes. Despite these localized deviations, G-R2R largely preserves the dominant spatial structure of RNF degradation across the four peak-loss conditions.

Fig. 15(b) further shows that larger relative errors occur mainly in the low-RNF regime. At lower reference RNF values, the same absolute prediction deviation corresponds to a larger relative error. Outside these severely degraded nodes, most predictions remain concentrated at relatively low error levels. Taken together, the spatial error maps and error–RNF relationships show that large prediction errors remain sparse and localized during the peak-degradation stages, while the principal spatial pattern and overall severity structure of RNF degradation are largely preserved.

### 6.5. Cross-Region Reusability and Practical Implications

The practical value of cross-region generalization is particularly evident in data-constrained target regions. Region-specific flood–transportation surrogates still depend on local flood-simulation models and simulation-derived training data, which can be difficult to establish where detailed drainage-network information or other data required for local physical modeling are incomplete or unavailable. G-R2R instead reuses predictive relationships learned from multiple source regions and can be applied to an unseen target network without generating a target-region simulation-labeled training set or retraining the model. This capability extends rapid RNF forecasting to regional networks where developing a dedicated physical model and region-specific surrogate would otherwise be difficult.

Cross-region reuse is accompanied by substantial forecasting efficiency. Once trained, G-R2R generates future RNF without repeatedly executing the complete SWMM-WCA2D-NIPW workflow. In the present case study, a full-region forecast requires approximately 3 s, compared with approximately 8.9 min for the complete simulation and network-functionality workflow, representing an approximately two-order-of-magnitude reduction in runtime. The computationally intensive simulation, label generation, and model training can be completed offline, while rapid inference is retained during the time-sensitive period when rainfall hazards are approaching or unfolding. Section 5.6 further shows that this offline requirement is adjustable: using 80% of the available training samples yields a MAPE of 3.71% and an $R^2$ of 0.833 in the unseen LX region, compared with 2.33% and 0.855 using the full training set, indicating a practical trade-off between training volume and prediction performance. If higher-fidelity and more computationally demanding physical models are used for RNF label generation, the relative advantage of avoiding their repeated execution during forecasting would be expected to increase further.

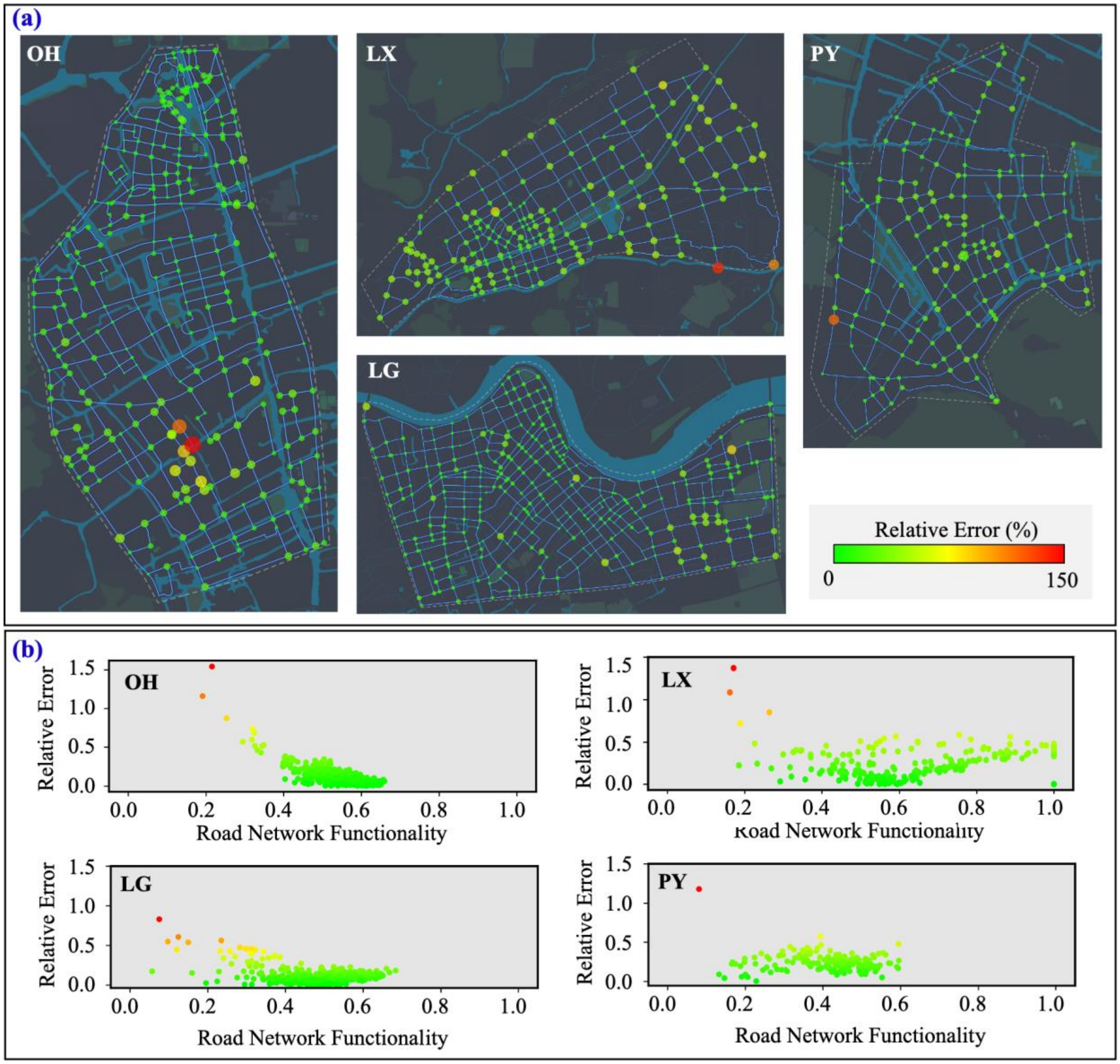

**Fig. 15. Peak-stage error characterization of G-R2R predictions across the four target regions.** (a) Node-level relative-error maps for OH, LX, LG, and PY at their respective peak RNF-loss stages. (b) Relationship between simulation-derived reference RNF and relative prediction error for each region.

The resulting benefit is the rapid provision of temporally and spatially resolved transportation-network functionality information in previously unseen regional networks. As shown in Sections 6.2-6.4, G-R2R captures the evolution and spatial concentration of network degradation and recovery, complementing local flood-condition information with network-level functional consequences. Such forecasts are therefore most relevant to network-level functional assessment, screening of substantially degraded areas, pre-positioning of emergency repair or support resources, and prioritization during network recovery.

## 7. Conclusions

This study developed G-R2R, a transferable spatiotemporal graph-learning framework for forecasting rainfall-induced road-network functionality across regional road networks. By integrating topology-flexible graph representation, region-conditioned static encoding, and dynamic rainfall–network feature learning, G-R2R learns shared predictive relationships from multiple source regions and applies them to previously unseen target networks without target-region retraining.

Experiments across ten study regions in Wenzhou support this capability within the regional and hydroclimatic conditions represented in this study. Under the benchmark setting, with Lingxi (LX) used as the unseen target region, G-R2R achieved a MAPE of 2.33% and an $R^2$ of 0.855. Across the ten leave-one-area-out experiments, in which each study region was held out in turn as the unseen target, MAPE values ranged from approximately 2.3% to 3.54%. Ablation experiments supported the contributions of the regional static representation and terrain-conditioned graph representation, while iterative forecasting showed moderate error growth over the evaluated 12-h rollout. In the Typhoon Fitow historical-event forecast replay, G-R2R reproduced the principal temporal trajectories and node-level spatial patterns of the simulation-derived RNF reference under observed rainfall forcing. A full-region forecast required approximately 3 s, compared with approximately 8.9 min for the complete flood-simulation and network-functionality workflow.

The cross-region capability is particularly relevant where detailed drainage-network information and local calibration records are incomplete or unavailable, as these data constraints can limit region-specific flood modeling, simulation-label generation, and surrogate development. G-R2R transfers predictive relationships learned from source regions and enables RNF forecasting in unseen target networks without target-region simulation-label generation or model retraining. Once trained, the model can rapidly update network-level functionality as rainfall conditions evolve. The resulting forecasts provide network-level information for assessing and screening substantial functionality degradation and, when interpreted together with normal-state network-performance information, can provide supporting information for emergency-resource pre-positioning and recovery prioritization.

Future research can extend the framework in several directions. The current prediction target, RNF, is a simulation-derived measure of network functionality. This formulation provides a consistent network-level prediction target for severe rainfall emergencies, for which systematic and continuous observations of transportation disruption remain scarce. Accordingly, the predictive accuracy reported in this study quantifies agreement with the simulation-derived RNF reference, while direct agreement with observed transportation disruption remains to be established. As richer extreme-event transportation observations become available, future studies can extend and evaluate the framework using directly observed measures such as road passability, traffic speed, and travel-time degradation. Such extensions should also account for event-induced changes in travel demand, which can alter how observed traffic conditions reflect underlying network disruption during severe events. Broader geographic evaluation is also needed because the current experiments are confined to regional road networks within the wider Wenzhou hydroclimatic setting. Cross-city and cross-climate studies should examine transfer under larger differences in rainfall regimes and flood-generation processes. These studies can further determine whether explicit climate-regime descriptors or adaptation mechanisms are needed to extend the current regional representation to hydroclimatically distinct domains. Evaluation using operational rainfall forecasts and explicit propagation of meteorological uncertainty would provide additional evidence for prospective forecasting applications.

## ACKNOWLEDGMENT

Zhejiang Department of Science and Technology (Grant No. 2024C03255)

## Appendix A. Regional calibration and validation of the flood-simulation models

### A.1. Regional calibration and validation procedure

The flood scenarios used for RNF label generation were produced using region-specific coupled SWMM–WCA2D models. SWMM was used to simulate surface runoff, drainage-network flow, and node overflow, while WCA2D was used to simulate the subsequent two-dimensional surface inundation process. Drainage-network configuration, catchment characteristics, terrain, and land-surface information were specified separately for each study region. The detailed SWMM–WCA2D implementation, drainage-network representation, and runoff-parameter settings follow Wang and Wang (2024).

For each of the ten study regions, the flood-simulation model was calibrated and validated using local drainage and flood-control information together with available inundation records. Model parameters were adjusted according to the corresponding drainage characteristics and documented inundation conditions before the calibrated models were used to generate the flood-simulation datasets. The calibration and validation procedure follows the regional modeling strategy adopted in Wang and Wang (2024).

LX is presented below as a representative case because detailed drainage-network, flood-control, and inundation information is available for this region.

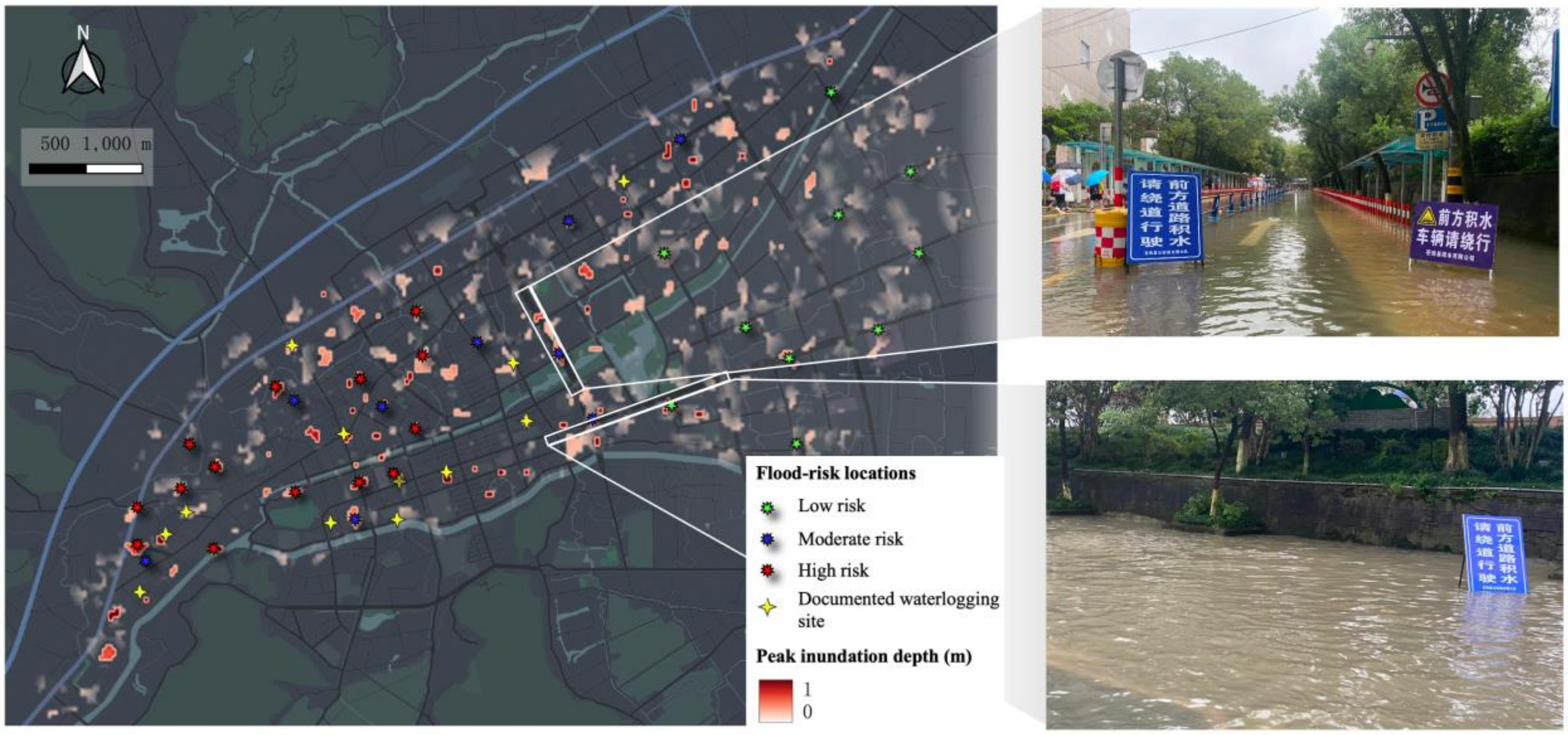

**Fig. A1.** Calibration and validation of the coupled SWMM–WCA2D flood model in LX under the 20-year return-period rainfall condition. The simulated peak-inundation depth is compared with reference flood-risk locations derived from high-resolution local simulations and documented waterlogging locations and inundated road sections identified from field photographs and reports. The simulated inundated area covers 93.3% of the combined reference locations.

### A.2. Representative calibration and validation in Lingxi

The coupled SWMM–WCA2D model for LX was calibrated under the 20-year return-period design rainfall condition using regional drainage and flood-control information. Following calibration, the simulated peak-inundation pattern was compared with two types of reference information. The first consisted of flood-risk locations derived from high-resolution simulations provided by the Cangnan County Housing and Urban–Rural Development Bureau under the same 20-year rainfall condition. The second consisted of 12 documented waterlogging locations and inundated road sections identified from field photographs and reports collected from local news and social-media records.

As shown in Fig. A1, the simulated peak-inundation pattern is spatially consistent with the reference flood-risk and documented waterlogging locations. Spatial coverage was calculated as the percentage of reference locations falling within the simulated inundation extent. The resulting spatial coverage reached 93.3%. The comparison shows that the calibrated SWMM–WCA2D model reproduces the principal spatial distribution of inundation-prone areas in Lingxi and provides the flood-depth fields subsequently used for road-network functionality assessment.